\documentclass[10pt,twocolumn,letterpaper]{article}

\usepackage{cvpr}              
\usepackage{chen}
\newcommand{\cmark}{\ding{51}}%
\usepackage[pagebackref,breaklinks,colorlinks]{hyperref}
\usepackage{soul}

\definecolor{mygray2}{gray}{.6}
\definecolor{mywarning}{RGB}{233,144,61}
\definecolor{myred}{RGB}{153,0,0}
\definecolor{myblue}{RGB}{0,31,95}
\definecolor{myorange}{RGB}{243,225,214}
\usepackage{amsthm}
\theoremstyle{definition}
\newtheorem{definition}{Definition}[subsection]

\newcommand{\pub}[1]{\color{gray}{\tiny{[{#1}]}}}

\usepackage{tabulary}
\newcolumntype{y}[1]{>{\raggedright\arraybackslash}p{#1pt}}
\newcolumntype{z}[1]{>{\raggedleft\arraybackslash}p{#1pt}}

\def\cvprPaperID{8609}
\def\confName{CVPR}
\def\confYear{2022}

\begin{document}

\title{Deep Hierarchical Semantic Segmentation}

\author{\!\!\!\!\!\!\!Liulei Li$^{1,5}$\thanks{Work done during an internship at Baidu Research.}~, Tianfei Zhou$^{2}$~, Wenguan Wang$^{3}$\thanks{Corresponding author: \textit{Wenguan Wang}.}~, Jianwu Li$^{1}$~, Yi Yang$^{4}$\\
\small{\!\!\!\!\!\!\!\!$^1$ Beijing Institute of Technology}~\small{$^2$ ETH Zurich}~\small{$^3$ ReLER, AAII, University of Technology Sydney}~\small{$^4$ CCAI, Zhejiang University}~\small{$^5$ Baidu Research}\\
\small\url{https://github.com/0liliulei/HieraSeg}
}

\maketitle

\begin{abstract}
Humans are able to recognize structured relations in observation, allowing us to decompose complex scenes into simpler parts and abstract the visual world in multiple levels. However, such hierarchical reasoning ability of human perception remains largely unexplored in current literature of semantic segmentation. Existing work is often aware of flatten labels and predicts target classes exclusively for each pixel. In this paper, we instead address hierarchical semantic segmentation (HSS), which aims at structured, pixel-wise description of visual observation in terms of a class hierarchy.  We devise \textsc{Hssn}, a general HSS framework that tackles two critical issues in this task: \textbf{i)} how to efficiently adapt existing hierarchy-agnostic segmentation networks to the HSS setting, and \textbf{ii)} how to leverage the hierarchy information~to regularize HSS network learning. To address \textbf{i)}, \textsc{Hssn} dire- ctly casts HSS as a pixel-wise multi-label classification task, only bringing minimal architecture change to current seg- mentation$_{\!}$ models.$_{\!}$ To$_{\!}$ solve$_{\!}$ \textbf{ii)},$_{\!}$ \textsc{Hssn}$_{\!}$ first$_{\!}$ explores inherent properties of the hierarchy as a training objective, which~en-\\
\noindent forces$_{\!}$ segmentation$_{\!}$ predictions$_{\!}$ to$_{\!}$ obey$_{\!}$ the$_{\!}$ hierarchy$_{\!}$ stru- cture. Further, with hierarchy-induced margin constraints, \textsc{Hssn}$_{\!}$ reshapes$_{\!}$ the$_{\!}$ pixel$_{\!}$ embedding$_{\!}$ space,$_{\!}$ so$_{\!}$ as$_{\!}$ to$_{\!}$ generate\\
\noindent  well-structured pixel representations and improve segmentation $_{\!}$eventually.$_{\!}$ We$_{\!}$ conduct$_{\!}$ experiments$_{\!}$ on$_{\!}$ four$_{\!}$ seman- tic segmentation datasets (\ie, Mapillary Vistas 2.0, City- scapes, LIP, and
PASCAL-Person-Part), with different class hierarchies, segmentation network architectures and backbones, showing the generalization and superiority of \textsc{Hssn}.
\end{abstract}

\vspace{-10pt}
\section{Introduction}
\begin{figure}[t]
	\begin{center}
		\includegraphics[width=\linewidth]{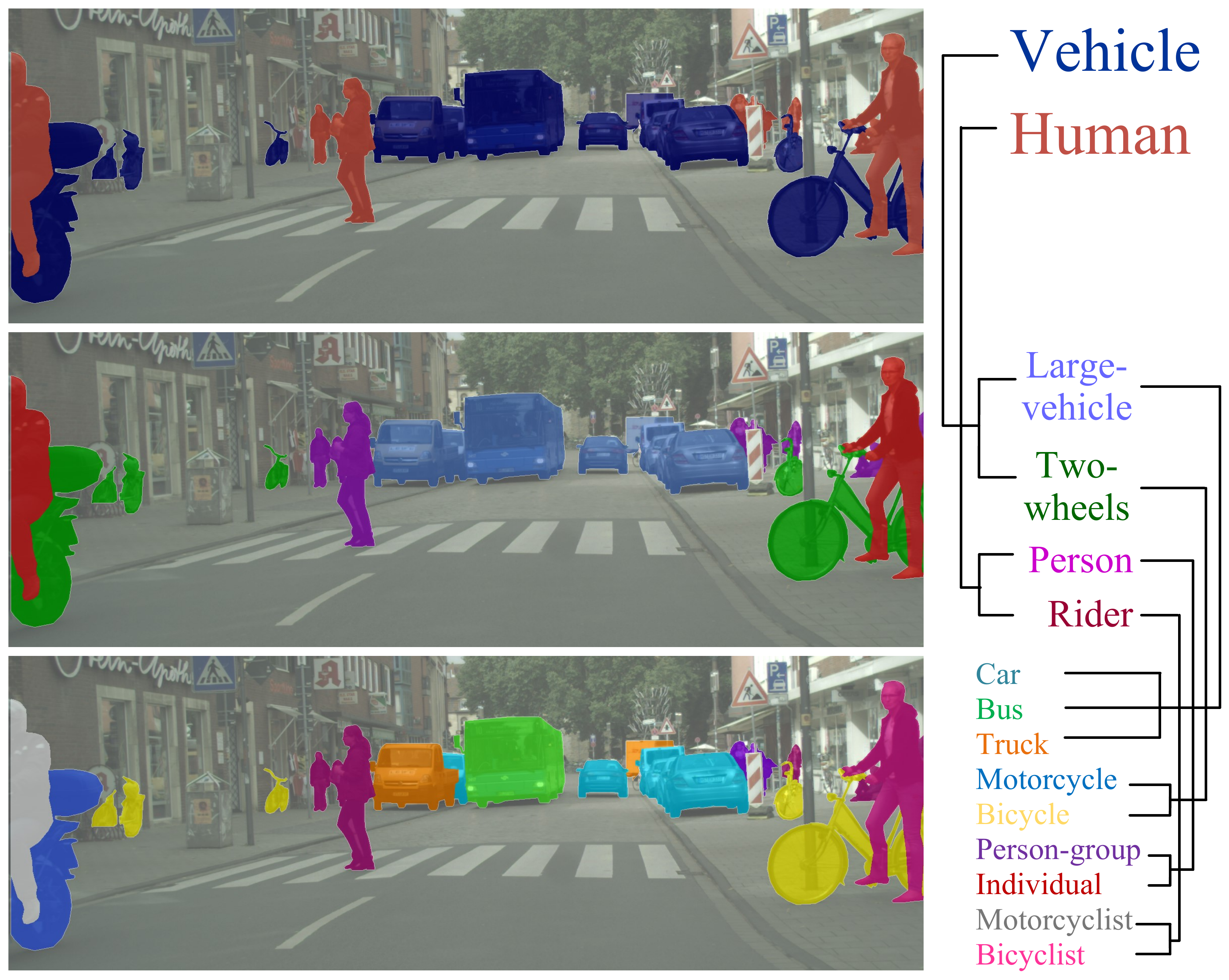}
	\end{center}
	\vspace{-18pt}
	\captionsetup{font=small}
	\caption{\small\!\!\textbf{Hierarchical semantic segmentation} explains visual scenes with multi-level abstraction (\textit{left}), by considering structured class relations (\textit{right}). The class taxonomy is borrowed from~\cite{neuhold2017mapillary}. }
	\vspace{-17pt}
	\label{fig:motivation}
\end{figure}

\vspace{-3pt}
Semantic segmentation, which aims to identify semantic categories for pixel observations, is viewed as a vital~step towards intelligent scene understanding~\cite{wang2021exploring}. The vast majority of modern segmentation models simply assume that~all the$_{\!}$  target$_{\!}$  classes$_{\!}$  are$_{\!}$  disjoint$_{\!}$  and$_{\!}$  should$_{\!}$  be$_{\!}$  distinguished$_{\!}$~ex- clusively$_{\!}$ during$_{\!}$ pixel-wise$_{\!}$ prediction.$_{\!}$ This$_{\!}$ fails$_{\!}$ to$_{\!}$ capture the$_{\!}$ structured$_{\!}$ nature$_{\!}$ of$_{\!}$ the$_{\!}$ visual$_{\!}$ world$_{\!}$~\cite{lu2021segmenting}:$_{\!}$ complex$_{\!}$ scenes arise from the composition of
simpler entities. Walking city, vehicles and pedestrian fill our view (Fig.~\ref{fig:motivation}). After focusing on the vehicles, we identify cars, buses, and trucks, which consist of more fine-grained parts like wheel and window. On the other hand, structured understanding of our world in terms of relations and  hierarchies is a central ability in human cognition~\cite{spelke2007core,yang2021multiple}. We group {chair} and {bed} as furniture, while {cat} and {dog} as {pet}. We understand this world over multiple levels of abstraction,  in order to maintain stable, coherent percepts in the face of complex visual inputs~\cite{kaiser2019object}.
The ubiquity of hierarchical decomposition serves as a core motivation behind many
structured machine learning models~\cite{dekel2004large,wehrmann2018hierarchical}, which have shown wide success in document classification~\cite{koller1997hierarchically,mccallum1998improving} and protein function prediction~\cite{valentini2010true,bi2011multilabel}.

In semantic segmentation literature,  surprisingly little~is understood about how to accommodate pixel recognition into semantic hierarchies. 
\cite{wang2019learning,wang2020hierarchical,xiao2018unified,liang2018dynamic,meletis2018training,li2020deep,wang2021hierarchical} are rare exceptions that exploit class hierarchies in segmentation networks. Nevertheless, they either focus specifically on the structured organization of human body parts~\cite{wang2019learning,wang2020hierarchical,wang2021hierarchical}, or introduce hierarchy-induced architectural changes to the segmentation network~\cite{xiao2018unified,liang2018dynamic,meletis2018training,li2020deep}, both hindering generality. More essentially, these methods are more aware of making efficient information propagation over the hierarchies (\eg, graph message passing~\cite{wang2020hierarchical,li2020deep,zhou2021cascaded}, multi-task learning~\cite{xiao2018unified}), without imposing tree-structured label dependencies/constraints into prediction and learning.

To mimic human hierarchical visual perception, we propose a novel approach for \textit{hierarchical semantic segmentation} (HSS).  In HSS, classes are not arranged in a ``flat'' structure, but organized as a tree-shaped hierarchy. Thus each pixel observation is associated to a root-to-leaf path of the class hierarchy (\eg, \texttt{human}$\rightarrow$\texttt{rider}$\rightarrow$\texttt{bicyclist}), capturing general-to-specific relations between classes. Our algorithm, called \textsc{Hssn}, addresses two core issues in HSS, yet untouched before. \textbf{First}, instead of previous structured segmentation models focusing on sophisticated network design, \textsc{Hssn} directly formulates HSS as a pixel-wise multi-label classification task. This allows to easily adapt existing segmentation models to the HSS setting, densely linking the fields of classic hierarchy-agnostic segmentation and HSS together. \textbf{Second}, \textsc{Hssn} makes full use of the class hierarchy in HSS network learning. To make pixel predictions coherent with the class hierarchy, \textsc{Hssn} explores two \textit{hierarchy constraints}, \ie, \textbf{i)} a pixel sample belonging to a given class must also belong to all its ancestors in the hierarchy, \textbf{ii)} a pixel sample not belonging to a given class must also not belong to all its descendants, as optimization criterion. This leads to a \textit{pixel-wise hierarchical segmentation learning} strategy, which enforces segmentation predictions to obey the hierarchy structure during training. \textsc{Hssn} further encodes the structured knowledge introduced by the class hierarchy into the pixel embedding space. This leads to a \textit{pixel-wise hierarchical representation learning} strategy, which inspires tree-induced margin separation for embedding space reshaping. As the hierarchy characterizes the underlying relationships between classes, \textsc{Hssn} is able to enrich pixel embeddings by pulling semantically similar pixels (\eg, \texttt{bicycle} and \texttt{motorcycle}) closer, while pushing semantically dissimilar  pixels (\eg, \texttt{pedestrian} and \texttt{lamppost}) farther away. This leads to more efficient learning by discovering and reusing common patterns \cite{garnot2020leveraging}, facilitating hierarchical segmentation eventually. This also allows our model to take different levels of mistakes into consideration. This is essential for some critical systems~\cite{bertinetto2020making}. Take autonomous driving as an example: mistaking a
\texttt{bicycle} for a \texttt{motorcycle} is less of a problem than confusing a \texttt{pedestrian} with a \texttt{lamppost}.

This work represents a solid step towards HSS. Our approach is elegant and principle; it is readily incorporated to arbitrary previous hierarchy-agnostic segmentation networks, with only marginal modification on the segmentation head. We train and test \textsc{Hssn} over four public benchmarks (\ie, Mapillary Vistas 2.0 \cite{neuhold2017mapillary}, Cityscapes \cite{cordts2016cityscapes},  LIP \cite{liang2018look}, PASCAL-Person-Part \cite{xia2017joint}), with different class hierarchies for urban street scene parsing and human semantic parsing. Extensive experimental results with different segmentation network architectures (\ie, DeepLabV3+$_{\!}$~\cite{chen2018encoder}, OCRNet~\cite{yuan2020object}, MaskFormer$_{\!}$~\cite{cheng2021maskformer}) and  backbones (\ie, ResNet-101$_{\!}$~\cite{he2016deep}, HRNetV2-W48$_{\!}$~\cite{wang2020deep}, Swin-Small$_{\!}$~\cite{liu2021swin}) verify the generalization and effectiveness of \textsc{Hssn}.

\section{Related Work}\label{sec:rw}

\noindent\textbf{(Hierarchy-Agnostic) Semantic Segmentation.} Semantic segmentation is to partition an image into regions with different semantic categories, which can be viewed as a pixel-wise classification task. Typical solutions for semantic segmentation follow a \textit{hierarchy-agnostic} setting, where each pixel is assigned to a single label from a set of disjoint semantic categories. In 2015, Long \etal proposed fully con-

\noindent volutional networks (FCNs)~\cite{long2015fully}, which are advantageous in end-to-end dense representation modeling, laying the foundation for modern semantic segmentation algorithms. As FCNs suffer from limited visual context with local receptive fields, how to effectively capture cross-pixel relations became the main focus of follow-up studies. Scholars devised many promising solutions, by enlarging receptive$_{\!}$ fields \cite{zhao2017pyramid,dai2017deformable,yang2018denseaspp,yu2015multi,chen2017deeplab,chen2018encoder},
building image pyramids \cite{lin2017refinenet,he2019adaptive}, exploring encoder-decoder architectures \cite{ronneberger2015u,chen2018encoder,badrinarayanan2017segnet}, utilizing boundary clues \cite{ding2019boundary,li2020improving,yuan2020segfix}, or incorporating$_{\!}$ neural$_{\!}$ attention$_{\!}$ \cite{harley2017segmentation,wang2018non,li2018pyramid,zhao2018psanet,sun2020mining,zhu2019asymmetric,li2019expectation,fu2019dual,huang2019ccnet,zhou2021group}.
Recently, a new family of semantic segmentation models \cite{xie2021segformer,strudel2021segmenter,zheng2021rethinking,cheng2021maskformer}, built upon the full attention (Transformer~\cite{vaswani2017attention}) architecture, yielded impressive performance, as it overcomes the issues in long-range cross-pixel dependency modeling.

Though impressive, existing semantic segmentation solutions rarely explore the structures between semantic concepts. We take a further step towards class relation aware semantic segmentation, which better reflects the structured nature of our visual world, and echoes the hierarchical reasoning mode of human visual perception. An appealing advantage of our  hierarchical solution is that, it can adapt existing class hierarchy-agnostic segmentation architectures, no matter FCN-based or Transformer-like, to the structured setting, in a simple and cheap manner.

\noindent\textbf{Scene$_{\!}$ Parsing/Hierarchical$_{\!}$ Semantic$_{\!}$ Segmentation.$_{\!}$} Our work is, at a high level, relevant to classical \textit{image parsing} algorithms~\cite{tu2005image,sudderth2005learning,sudderth2008describing,han2008bottom,yao2012describing}. Image parsing has been extensively studied in the pre-deep learning era, dating back to~\cite{tu2005image}. Image parsing seeks a  \textit{parse graph} that explains visual observation following a ``divide-and-conquer''strategy:

\noindent a football game image is first parsed into person, sports field, and spectator, which are further decomposed, \eg, person consists of face and body patterns. In the deep learning era, \textit{human parsing}, as a sub-field of scene parsing, became active. Some recent human parsers explored human part relations, based on the human hierarchy \cite{zhu2018progressive,wang2019learning,wang2020hierarchical,ji2020learning,meletis2018training,zhou2021differentiable}. Only very few efforts~\cite{zhao2017open,xiao2018unified,liang2018dynamic,li2020deep} are concerned with utilizing structured knowledge to aid the training of general-purpose semantic segmentation networks.

To accommodate the semantic structures imposed by the hierarchy, previous methods tend to greatly change the segmentation network, through the use of different graph neural networks. They hence put all emphasis on how to~aggre- gate information over the structured network. Beyond their specific solutions, we propose a general framework for both HSS network design and training. This leads to an elegant view of how to adapt typical segmentation networks to the class hierarchy with only minimal architecture change, and how to involve the hierarchy for regularizing network training, which are core problems yet ignored by prior methods.

\noindent\textbf{Hierarchical Classification.} Considering class hierarchies when designing classifiers is a common issue across various machine learning application domains~\cite{silla2011survey}, such as text categorization~\cite{rousu2006kernel}, functional genomics~\cite{barutcuoglu2006hierarchical}, and image classification~\cite{deng2010does,bengio2010label}. Depending on whether each datapoint can be assigned a single path or multiple paths in the hierarchy, hierarchy-aware classification can be categorized into \textit{hierarchical classification}~\cite{koller1997hierarchically,mccallum1998improving,dekel2004large,sun2001hierarchical} and, a more general setting, \textit{hierarchical multi-label classification}~\cite{bi2011multilabel,wehrmann2018hierarchical,giunchiglia2020coherent}. In the field of computer vision, exiting efforts for class taxonomy aware image classification can be broadly divided into three groups~\cite{bertinetto2020making}: i) \textit{Label-embedding methods}~\cite{bengio2010label,frome2013devise,akata2015evaluation,xian2016latent} that embed class labels to vectors whose relative locations represent semantic relationships; ii) \textit{Hierarchical losses}~\cite{deng2010does,zhao2011large,verma2012learning,bilal2017convolutional,bertinetto2020making} which are designed to inspire the coherence between the prediction and class hierarchy; and iii) \textit{Hierarchical architectures}~\cite{zweig2007exploiting,yan2015hd,ahmed2016network,zhu2017b} that adapt the classifier architecture to the class hierarchy.

Drawing inspiration from these past efforts, we advocate for holistic visual scene understanding through pixel-level hierarchical reasoning. We leverage tree-structured class dependencies as supervision signal to guide hierarchy-coherent pixel prediction and structured pixel embedding.

\noindent\textbf{Hierarchical Embedding.}  The objective of an embedding algorithm is to organize data samples (\eg, words, images) into an high-dimensional space where their distance reflects
their semantic similarity~\cite{nickel2017poincare}. As semantics are inherently structured,  it is necessary to integrate different levels of concept abstraction into representation embedding. Some algorithms directly parameterize the hierarchical embedding space into hierarchical models \cite{weinberger2009large,verma2012learning,niu2017hierarchical,mousavi2017hierarchical,chen2018fine,yang2020hierarchical}. While straightforward, they are computationally intensive
and have to adjust the network architecture when handling different hierarchies. Some alternatives \cite{ge2018deep,yang2019adaptive,barz2019hierarchy,kerola2021hierarchical} design hierarchy-aware metric learning objectives~\cite{nickel2017poincare,ganea2018hyperbolic,sala2018representation} to directly shape the embedding space.

With a similar spirit, in this work, we adopt semantic hierarchy-induced margin separation to reinforce pixel representation learning and make prediction less ambiguous.

\section{Our Approach}\label{sec:method}
	\vspace{-2pt}
Our goal is to accommodate standard semantic segmentation networks to the HSS problem and then exploit structured class relations in order to generate hierarchy-coherent representations and predictions,~and improve performance.$_{\!}$ Given this goal, we develop$_{\!}$ \textsc{Hierarchical$_{\!}$ Semantic$_{\!}$ Segmentation$_{\!}$ Networks$_{\!}$} (\textsc{Hssn}), a general framework for HSS network design (\S\ref{sec:hsss}) and training  (\S\ref{sec:hasl}).
	\vspace{-4pt}
\subsection{Hierarchical Semantic Segmentation Networks}\label{sec:hsss}
	\vspace{-2pt}
Rather than typical segmentation methods treating semantic classes as disjoint labels, in the HSS setting, the underlying dependencies between classes are considered and formalized in a form of a tree-structured hierarchy, $\mathcal{T}_{\!}\!=_{\!}\!(\mathcal{V}, \mathcal{E})$. Each node $v_{\!}\!\in_{\!}\!\mathcal{V}_{\!}$ denotes a semantic class/concept, while each edge $(u, v)_{\!}\!\in_{\!}\!\mathcal{E}_{\!}$ encodes the decomposition relationship between two classes, $u, v\!\in\!\mathcal{V}$, \ie, parent node $v$ is a more general, superclass of child node $u$, such as $(u, v)\!=\!(\texttt{bicycle},\texttt{vehicle})$. We assume $(v, v)\!\in\!\mathcal{E}$,~thus every class is both a subclass and superclass of itself.  The root node of $\mathcal{T}$, \ie, $v^r$, denotes the most general class. The leaf nodes, \ie,  $\mathcal{V}_{\chi}$, refer to the most fine-grained classes, such as $\mathcal{V}_{\chi\!}\!=\!\{\texttt{tree}, \texttt{bicyclist}, \cdots\}$ in urban street scene parsing, and $\mathcal{V}_{\chi\!}\!=\!\{\texttt{head}, \texttt{leg}, \cdots\}$ in human parsing.

For a typical hierarchy-agnostic segmentation network, an encoder $f_{\text{ENC\!}}$ is first adopted to map an image $I$ into a dense feature tensor $\bm{I}\!=\!f_{\text{ENC}}(I)\!\in\!\mathbb{R}^{H\!\times\!W\!\times\!C}$, where $\bm{i}\!\in\!\bm{I}$ is

\noindent the embedding of pixel $i_{\!}\!\in_{\!}\!I$. Then a segmentation head $f_{\text{SEG}\!}$ is used to get a score map $\bm{Y}\!=\!\texttt{softmax}(f_{\text{SEG}}(\bm{I}))\!\in\![0,1]^{H\!\times\!W\!\times\!|\mathcal{V}_{\chi\!}|}$$_{\!}$  \wrt \textbf{the leaf node set} $\mathcal{V}_{\chi}$. Given the \textit{score vector} ${\bm{y}}_{\!}\!=_{\!}\![y_{{v}_{\chi\!}}]_{{v}_{\chi\!}\in\mathcal{V}_{\chi\!\!}}\!\in_{\!}\![0,1]^{|\mathcal{V}_{\chi\!}|\!}$ and \textit{groundtruth leaf label} $\hat{v}_{\chi\!}\!\in\!\!\mathcal{V}_{\chi\!}$ for pixel $i$,  the categorical cross-entropy loss is optimized:\!\!
\vspace{-1pt}
\begin{equation}\label{eq:CCE}
\begin{aligned}\small
\mathcal{L}^{\text{CCE}}(\bm{y})\!=\!-\log(y_{\hat{v}_{\chi}}).
	\end{aligned}
\vspace{-0pt}
\end{equation}
During inference, pixel $i$ is associated to \textit{a single leaf node}: $v^*_{\chi}\!=\!\argmax_{v_{\chi}}(y_{{v}_{\chi}})$.

To accommodate classic segmentation networks to the HSS setting with minimum change, our \textsc{Hssn} first formulates HSS as a pixel-wise multi-label classification task, \ie, map pixels with their corresponding classes in the hierarchy as a whole. Specifically, \textit{only} the segmentation head $f_{\text{SEG}\!}$ is modified to predict an \textit{augmented} score map $\bm{S}\!=\!\texttt{sigmoid}(f_{\text{SEG}}(\bm{I}))\!\in\![0,1]^{H\!\times\!W\!\times\!|\mathcal{V}|}$ \wrt \textbf{the entire class hierarchy} $\mathcal{V}$. Given the {score vector} ${\bm{s}}\!=\![s_{v}]_{{v}\in\mathcal{V}\!}\!\in\![0,1]^{|\mathcal{V}|\!}$ and \textit{groundtruth binary label set}~$\hat{\bm{l}}\!=\![\hat{l}_{v}]_{{v}\in\mathcal{V}\!}\!\in\!\{0,1\}^{|\mathcal{V}|\!}$ for pixel $i$,  the binary cross-entropy loss is optimized:\!\!
\vspace{-3pt}
\begin{equation}
	\begin{aligned}\small\label{eq:sigmoidbce}
		\mathcal{L}^{\text{BCE}}({\bm{s}}) = \textstyle\sum\nolimits_{v\in\mathcal{V}}\!-\hat{l}_v\!\log(s_{v}) \!-\! (1 \!-\! \hat{l}_v)\!\log(1\!-\!s_{v}).
	\end{aligned}
	\vspace{-1pt}
\end{equation}
During inference, each pixel $i$ is associated with the top-scoring root-to-leaf path in the class hierarchy $\mathcal{T}$:
\vspace{-3pt}
\begin{equation}
	\begin{aligned}\small\label{eq:infer}
		\{v^*_1, \cdots, v^*_{|\mathcal{P}|}\}=\!\argmax_{\mathcal{P}\subseteq\mathcal{T}}\textstyle\sum\nolimits_{v_p\in\mathcal{P}}s_{v_p},
	\end{aligned}
	\vspace{-3pt}
\end{equation}
where $\mathcal{P}\!=\!\{v_1, \cdots, v_{|\mathcal{P}|}\}\!\subseteq\!\mathcal{T}$ denotes a feasible root-to-leaf path of $\mathcal{T}$, \ie, $v_{1\!}\!\in\!\mathcal{V}_{\chi}$, $v_{|\mathcal{P}|\!}\!=\!v^r$, and $\forall v_{p}, v_{p+1\!}\!\in\!\mathcal{P}\!\Rightarrow\!(v_{p}, v_{p+1})\!\in\!\mathcal{E}$. Although Eq.~\ref{eq:infer} ensures the coherence between pixel-wise prediction and the class hierarchy during the inference stage, there is no any class relation information used for segmentation network training, as the binary cross-entropy loss in Eq.~\ref{eq:sigmoidbce} is computed over each class independently. To alleviate this issue, we propose a hierarchy-aware segmentation learning scheme (\S\ref{sec:hasl}), which incorporates the semantic structures into the training of \textsc{Hssn}.

\begin{figure}[t]
	\vspace{-12pt}
	\begin{center}
		\includegraphics[width=\linewidth]{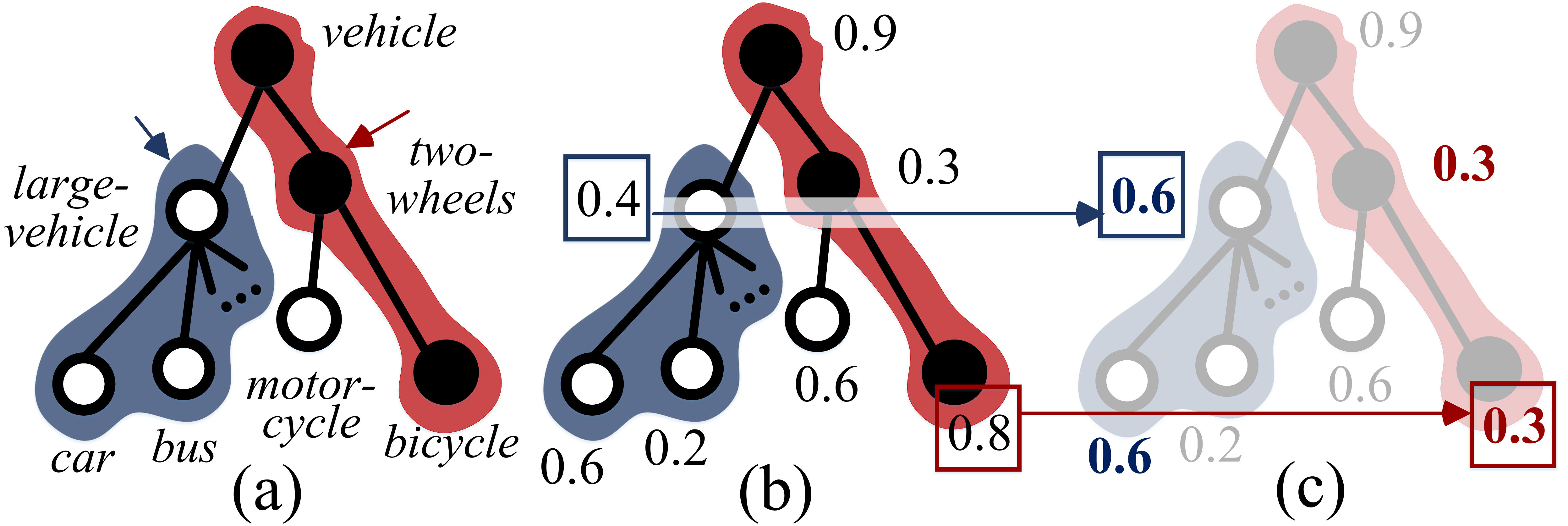}
		\put(-250,64.5){\tiny \textbf{\textcolor[RGB]{91,105,134}{Negative $\mathcal{T}$\!-Property}}}
		\put(-185.1,65.5){\tiny \textbf{\textcolor[RGB]{189,91,83}{Positive $\mathcal{T}$\!-Property}}}
	\end{center}
	\vspace{-18pt}
	\captionsetup{font=small}
	\caption{\small\textbf{Hierarchy constraints} used in our pixel-wise hierarchical segmentation learning (\S\ref{sec:hsl}). (a) In the class hierarchy, the filled circles represent the positive classes, while empty circles indicate the negative classes. The positive and negative $\mathcal{T}$-properties are highlighted in the red and blue regions, respectively. (b) The original score vector $\bm{s}$ predicted for the class hierarchy. The predictions which violate the positive and negative $\mathcal{T}$-constraints are highlighted in the red and blue rectangles, respectively. (c) The updated score vector  $\bm{p}$, which satisfies the $\mathcal{T}$-constraints. With $\mathcal{L}^{\text{TM}}$, the penalties for the wrong predictions, \ie, `{\color{myblue}\textbf{0.6}}' and `{\color{myred}\textbf{0.3}}', are increased twice, compared with applying $\mathcal{L}^{\text{BCE}}$ on (b).}
	\vspace{-12pt}
	\label{fig:fig2}
\end{figure}	

\vspace{-3pt}
\subsection{Hierarchy-Aware Segmentation Learning}\label{sec:hasl}
	\vspace{-3pt}
Our hierarchy-aware segmentation learning scheme includes two major components: i) a \textit{pixel-wise hierarchical segmentation learning} strategy (\S\ref{sec:hsl}) which supervises the segmentation prediction $\bm{S}$ in a hierarchy-coherent manner, and ii) a \textit{pixel-wise hierarchical representation learning}  strategy (\S\ref{sec:hrl}) that makes hierarchy-induced margin separation for reshaping the pixel embedding space $f_{\text{ENC}}$.
	\vspace{-8pt}
\subsubsection{Pixel-Wise Hierarchical Segmentation Learning}\label{sec:hsl}
	\vspace{-3pt}
For each pixel, the assigned labels are hierarchically consistent if they satisfy the following two properties (Fig.\!~\ref{fig:fig2}):
\vspace{-15pt}
\begin{definition}[Positive $\mathcal{T}$-Property] \label{definition:PP}
\textit{For each pixel, if a class is labeled positive, all its ancestor nodes (\ie, superclasses) in $\mathcal{T}$ should be labeled positive.}
\vspace{-5pt}
\end{definition}

\vspace{-5pt}
\begin{definition}[Negative $\mathcal{T}$-Property] \label{definition:NP}
\textit{For each pixel, if a class is labeled negative, all its child nodes (\ie, subclasses) in $\mathcal{T}$ should be labeled negative.}
\vspace{-4pt}
\end{definition}

The first property, also known as $\mathcal{T}$-property~\cite{bi2011multilabel}, was explored in some hierarchical classification work \cite{vendrov2015order,wehrmann2018hierarchical,giunchiglia2020coherent},  while the second property is ignored. Actually, these two properties are complementary and crucial for consistent hierarchical prediction. Specifically, to incorporate these two label consistency properties into the supervision of \textsc{Hssn}, we further derive the following  two hierarchy constraints \wrt per-pixel prediction, \ie, ${\bm{s}}\!=\![s_{v}]_{{v}\in\mathcal{V}\!}\!\in\![0,1]^{|\mathcal{V}|}$:
\vspace{-3pt}
\begin{definition}[Positive $\mathcal{T}$-Constraint] \label{definition:PC}
\textit{For each pixel, if $v$ class is labeled positive, and $u$ is an ancestor node (\ie, superclass) of $v$, it should hold that $s_v\!\leq\!s_u$.}
\vspace{-5pt}
\end{definition}
\vspace{-6pt}
\begin{definition}[Negative $\mathcal{T}$-Constraint] \label{definition:NC}
\textit{For each pixel, if $v$ class  is labeled negative, and $u$ is a child node (\ie, subclass) of $v$, it should hold  that $1-s_v\!\leq\!1-s_u$.}
\vspace{-5pt}
\end{definition}

With the positive $\mathcal{T}$-constraint (\textit{cf}.~Def.~\ref{definition:PC}), the positive $\mathcal{T}$-property (\textit{cf}.~Def.~\ref{definition:PP}) can be always guaranteed. Similar conclusion is also hold for the negative $\mathcal{T}$-constraint (\textit{cf}.~Def.~\ref{definition:NC}) and negative $\mathcal{T}$-property (\textit{cf}.~Def.~\ref{definition:NP}).

\noindent\textbf{Tree-Min Loss.} To ensure the satisfaction of the two hierarchy constraints, we estimate a hierarchy-coherent score map $\bm{P}\!\in\![0,1]^{H\!\times\!W\!\times\!|\mathcal{V}|}$ from $\bm{S}$. For pixel $i$, the updated score vector $\bm{p}\!=\![p_{v}]_{{v}\in\mathcal{V}\!}\!\in\![0,1]^{|\mathcal{V}|}$ in $\bm{P}$ is given as:
\vspace{-3pt}
\begin{equation}
\begin{aligned}\label{eq:max}\small
\!\!\!\left\{
\begin{aligned}
p_v &\!=\min_{u\in\mathcal{A}_v} (s_u)~~~~~~~~~~~~~~~~~~~~~~~~~~~~~~~~~~~\!\text{if~~} \hat{l}_v\!=\!1,\!\\[-0.7ex]
1\!-\!p_v &\!=\!\min_{u\in\mathcal{C}_v} (1\!-\!s_u) \!=\! 1\!-\!\max_{u\in\mathcal{C}_v} (s_u)~~~~~~\!\text{if~~} \hat{l}_v\!=\!0,\!
\end{aligned}
\right.
\end{aligned}
	\vspace{-2pt}
\end{equation}
where $\mathcal{A}_v$ and $\mathcal{C}_{v\!}$ denote the superclass and subclass sets of $v$ in $\mathcal{T}$ respectively, and $\bm{s}\!=\![s_{v}]_{{v}\in\mathcal{V}\!}\!\in_{\!}\!\bm{S}$ refers to the original score vector of pixel $i$. Note that, according to our definition $(v,v)\in\mathcal{E}$ (\textit{cf}.~\S\ref{sec:hsss}), we have $v\!\in\!\mathcal{A}_v$ and $v\!\in\!\mathcal{C}_v$. With Eq.~\ref{eq:max}, the pixel-wise prediction $\bm{p}$ is guaranteed to always satisfy the hierarchy constraints (\textit{cf}.~Defs.~\ref{definition:PC} and \ref{definition:NC}).

We thus build a hierarchical segmentation training objective, \ie, tree-min loss, to replace $\mathcal{L}^{\text{BCE}}({\bm{s}})$ in Eq.~\ref{eq:sigmoidbce}:
\vspace{-2pt}
\begin{equation}\small
	\begin{aligned}\label{eq:TML}
		\!\!\!\mathcal{L}^{\text{TM}\!}(\bm{p}) &=\textstyle\sum\limits_{v\in\mathcal{V}}-\hat{l}_v\!\log(p_{v}) - (1 - \hat{l}_v)\!\log(1\!-\!p_{v}),\!\!\!\\
&=\textstyle\sum\limits_{v\in\mathcal{V}}-\hat{l}_v\!\log(\min\limits_{u\in\mathcal{A}_v}({s}_u)) - \\[-0.7ex]
&~~~~~~~~~~~~~(1 - \hat{l}_v)\!\log(1 -\max_{u\in\mathcal{C}_v} (s_u)).\!\!\!
	\end{aligned}
	\vspace{-2pt}
\end{equation}
Compared with $\mathcal{L}^{\text{BCE}}(\bm{s})$, $\mathcal{L}^{\text{TM}}(\bm{p})$ is more favored as the structured score distribution $\bm{p}$ is constructed by strictly following the hierarchy constraints (\textit{cf}.~Eq.~\ref{eq:max}), and
hence the violation of the hierarchy properties (\ie, any undesired prediction of $\bm{p})$ can be explicitly penalized (see Fig.\!~\ref{fig:fig2}(c)).

\begin{figure}[t]
	\vspace{-12pt}
	\begin{center}
		\includegraphics[width=\linewidth]{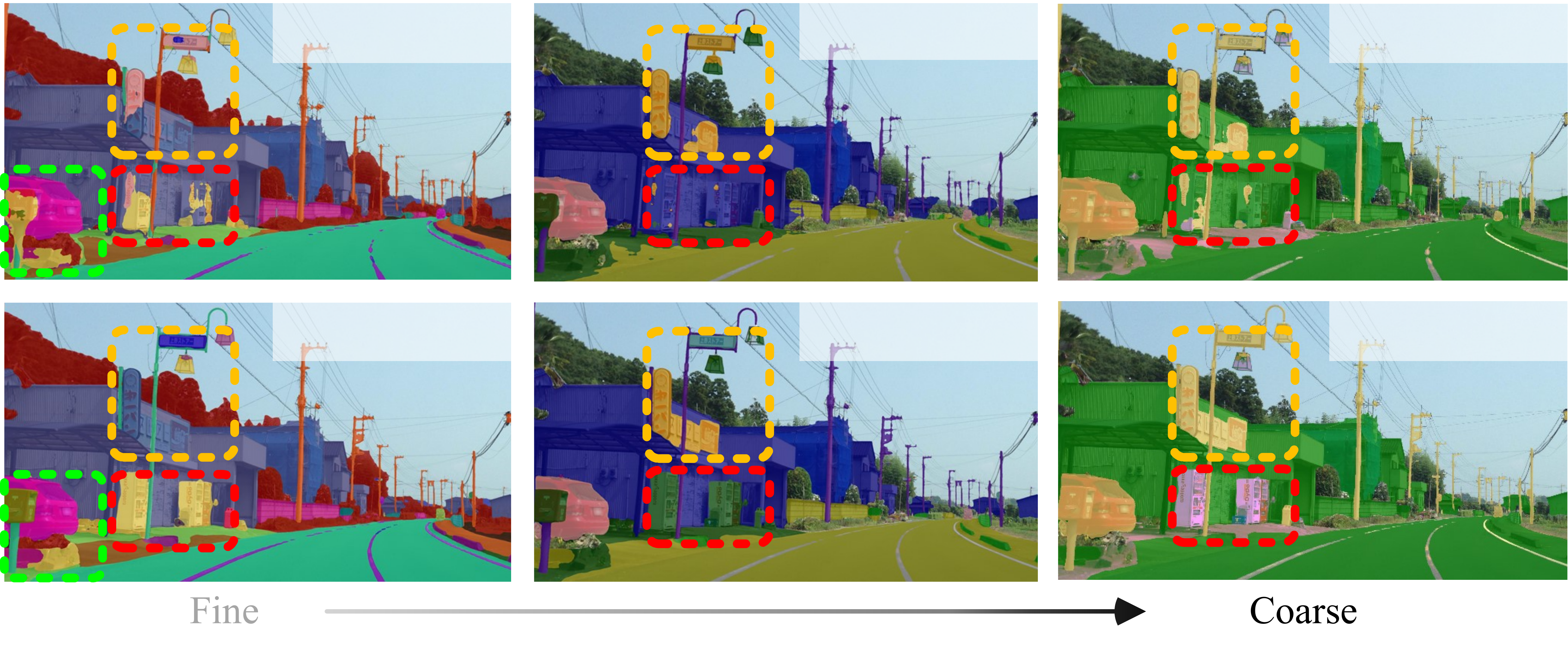}
		\put(-197.8,90.5){ \tiny $\text{mIoU}^1$: $\text{51.42\%}$ }
		\put(-197.8,45.2){ \tiny $\text{mIoU}^1$: $\text{72.83\%}$ }
		\put(-118.0,90.5){ \tiny $\text{mIoU}^2$: $\text{69.15\%}$ }
		\put(-118.0,45.2){ \tiny $\text{mIoU}^2$: $\text{85.61\%}$ }
		\put(-38.0,90.5){ \tiny $\text{mIoU}^3$: $\text{78.54\%}$ }
		\put(-38.0,45.2){ \tiny $\text{mIoU}^3$: $\text{92.37\%}$ }

	\end{center}
	\vspace{-18pt}
	\captionsetup{font=small}
	\caption{\small{Effect of $\mathcal{L}^{\text{BCE}}$ in Eq.~\ref{eq:sigmoidbce} (top) vs $\mathcal{L}^{\text{FTM}}$ in Eq.~\ref{eq:FocalTML} (bottom).}}
	\vspace{-12pt}
	\label{fig:3}
\end{figure}
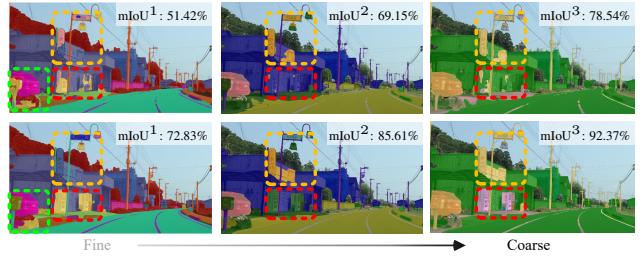	

\noindent\textbf{Focal Tree-Min Loss.} Inspired by the focal loss~\cite{lin2017focal}, we add a modulating factor to the tree-min loss (\textit{cf}.~Eq.~\ref{eq:TML}), so as to reduce
the relative loss for well-classified pixel samples and focus on
those difficult ones:
\begin{equation}\small
	\begin{aligned}\label{eq:FocalTML}
		\!\!\!\!\!\!\!\!\!\mathcal{L}^{\text{FTM}\!}(\bm{p}) &\!=\!\!\!\textstyle\sum\limits_{v\in\mathcal{V}}\!\!-\hat{l}_v(1\!-\!p_{v\!})^{\gamma\!}\log(p_{v\!})\!-\!(1\!-\! \hat{l}_{v\!})(p_{v\!})^{\gamma\!}\log(1\!-\!p_{v\!}),\!\!\!\!\!\!\\
		&\!=\!\!\!\textstyle\sum\limits_{v\in\mathcal{V}}\!\!-\hat{l}_v(1 \!-\!\min\limits_{u\in\mathcal{A}_v}({s}_u))^{\gamma\!}\log(\min\limits_{u\in\mathcal{A}_v}({s}_u)) - \!\!\!\\[-0.7ex]
		&~~~~~~~~(1\! - \!\hat{l}_v)(\max_{u\in\mathcal{C}_v} (s_u))^{\gamma\!}\log(1 \!-\!\max_{u\in\mathcal{C}_v} (s_u)),\!\!\!\!\!\!
	\end{aligned}
\end{equation}
where $\gamma\!\geq\!0$ is a tunable focusing parameter controling the rate at which easy classes are down-weighted. When $\gamma\!=\!0$, $\mathcal{L}^{\text{FTM}}(\bm{p})$ is equivalent to $\mathcal{L}^{\text{TM}}(\bm{p})$. Fig.~\ref{fig:3} shows representative visual effects of $\mathcal{L}^{\text{FTM}}$ against $\mathcal{L}^{\text{BCE}}$. We see that $\mathcal{L}^{\text{FTM}}$ yields more precise and coherent results. In \S\ref{sec:ablationstudy}, we provide quantitative comparison results for $\mathcal{L}^{\text{BCE}}(\bm{s})$ (\textit{cf}.~Eq.~\ref{eq:sigmoidbce}), $\mathcal{L}^{\text{TM}}(\bm{p})$ (\textit{cf}.~Eq.~\ref{eq:TML}), and $\mathcal{L}^{\text{FTM}}(\bm{p})$ (\textit{cf}.~Eq.~\ref{eq:FocalTML}).

	\vspace{-7pt}
\subsubsection{$_{\!}$Pixel-Wise Hierarchical Representation Learning$_{\!\!\!}$}\label{sec:hrl}
\vspace{-3pt}
Through mapping pixels with their corresponding semantic classes in the hierarchy $\mathcal{T}$ as a whole (\textit{cf}.~\S\ref{sec:hsss}), we exploit intrinsic properties of $\mathcal{T}$ (\textit{cf}.~Defs.~\ref{definition:PP}-\ref{definition:NP}) as constraints (\textit{cf}.~Defs. \ref{definition:PC}-\ref{definition:NC}) to encourage hierarchy-coherent segmentation prediction $\bm{S}$ (\textit{cf}.~Eqs.~\ref{eq:TML}-\ref{eq:FocalTML}).  As the class hierarchy provides
rich semantic relations among categories over different levels of concept abstraction, next we will exploit such structured knowledge to reshape the pixel embedding space $f_{\text{ENC}}$, so as to generate more efficient pixel representations and improve final segmentation performance.

With this purpose, we put forward a margin based pixel-wise hierarchical representation learning strategy, where the learned pixel embeddings are well separated with structured margins imposed by the class hierarchy $\mathcal{T}$. Specifically, for any pair of labels $u,v\!\in\!\mathcal{V}$, let $\psi(u,v)$ denote their \textit{distance} in the tree $\mathcal{T}$. That is,  $\psi(u,v)$ is defined as the length (in edges) of the shortest path between $u$ and $v$ in $\mathcal{T}$. The distance function $\psi(\cdot,\cdot)$ is in fact a semantic similarity metric defined over $\mathcal{T}$~\cite{dekel2004large}; it is a non-negative and symmetric function, $\psi(v,v)\!=\!0$, $\psi(u,v)\!=\!\psi(v,u)$, and the triangle inequality always holds with equality.

In \textsc{Hssn},  the structured margin constraints are defined by the tree distance $\psi(\cdot,\cdot)$, leading to a \textbf{tree-triplet loss}. This loss is optimized on a set of pixel triplets $\{i^{ }, i^+, i^{-\!}\}$, where $i^{ }, i^{+}, i^{-\!}$ are anchor, positive and negative pixel samples, respectively.
$\{i^{}, i^+, i^-\}$ are sampled from the whole training batch, such that $\psi(\hat{v}^{}_{\chi},\hat{v}^+_{\chi})\!<\!\psi(\hat{v}^{}_{\chi},\hat{v}^-_{\chi})$, where $\hat{v}^{}_{\chi}$, $\hat{v}^+_{\chi}, \hat{v}^{-\!}_{\chi\!}\!\in\!\mathcal{V}_{\chi\!}$ are the groundtruth leaf labels of $i$, $i^+$, and $i^-$, respectively. As such, in our tree-triplet loss, the positive samples are more semantically similar to the anchor pixels (\ie, closer in $\mathcal{T}$), compared with the negative pixels. Note that this is different from the classic, hierarchy-agnostic triplet loss \cite{schroff2015facenet}, where the anchor and positive samples are from the same class, while the anchor and negative samples are from different classes, \ie, $\hat{v}^{}_{\chi}\!=\!\hat{v}^+_{\chi}$, and $\hat{v}^{}_{\chi}\!\neq\!\hat{v}^-_{\chi}$. With a valid training triplet $\{i^{}, i^+, i^{-\!}\}$, our loss is given as:
\vspace{-3pt}
\begin{equation}
	\begin{aligned}\label{eq:TTL}
		\mathcal{L}^{\text{TT}}(\bm{i}^{}, \bm{i}^{+}, \bm{i}^{-\!}) = \max\{\langle\bm{i}, \bm{i}^{+}\rangle - \langle\bm{i}, \bm{i}^{-}\rangle + m, 0\},
	\end{aligned}
	\vspace{-1pt}
\end{equation}
where $\bm{i}^{}, \bm{i}^{+}, \bm{i}^{-\!}\!\in\!\mathbb{R}^C$ are the embeddings of $i$, $i^+$, and $i^-$, respectively, obtained from the encoder $f_{\text{ENC}}$, $\langle\cdot,\cdot\rangle$ is a distance function to measure the similarity of two inputs; we use the cosine distance, \ie, $\langle \bm{x},\bm{y} \rangle\!=\!\frac{1}{2}(1\!-\!\frac{\bm{x}\cdot\bm{y}}{\|\bm{x}\|\|\bm{y}\|})\!\in\![0,1]$. The margin $m$ forces the gap of $\langle\bm{i}, \bm{i}^{-}\rangle$  and $\langle\bm{i}, \bm{i}^{+}\rangle$ larger than $m$. When the gap is larger than $m$, the loss value would be zero. The separation margin $m$ is determined as:
	\vspace{-4pt}
\begin{equation}
	\begin{aligned}\small\label{eq:margin}
		m &= m_\varepsilon + 0.5 m_\tau \\
m_\tau&=(\psi(\hat{v}^{}_{\chi},\hat{v}^{-\!}_{\chi}) - \psi(\hat{v}^{}_{\chi},\hat{v}^{+\!}_{\chi}))/2D ,
	\end{aligned}
	\vspace{-2pt}
\end{equation}
where $m_\varepsilon\!\!=\!\!0.1$ is set as a \textit{constant} for the tolerance of the intra-class variance, \ie, maximum intra-class distance, $m_\tau\!\in\![0,1]$ is a \textit{dynamic} violate margin, which is computed according to the semantic relationships among $i^{ }$, $i^{+}$, and $i^{-\!}$ over the class hierarchy $\mathcal{T}$, and  $D$ refers to the height of $\mathcal{T}$.

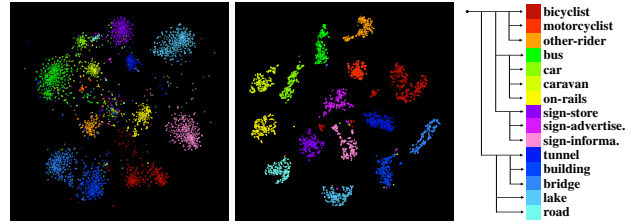
\begin{figure}[t]
	\begin{center}
		\begin{tabular}{ccc}
		\includegraphics[scale=0.26]{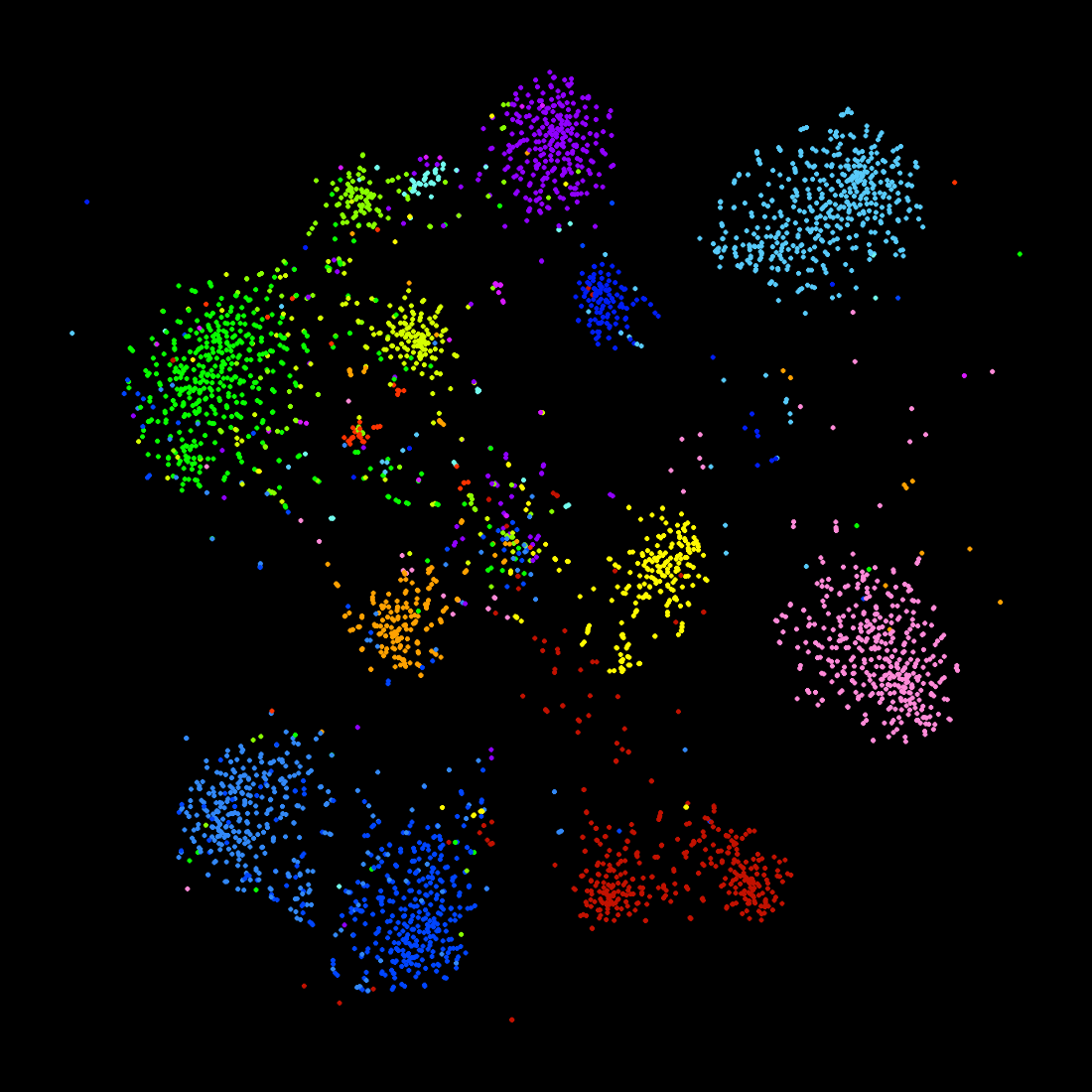}
		\includegraphics[scale=0.26]{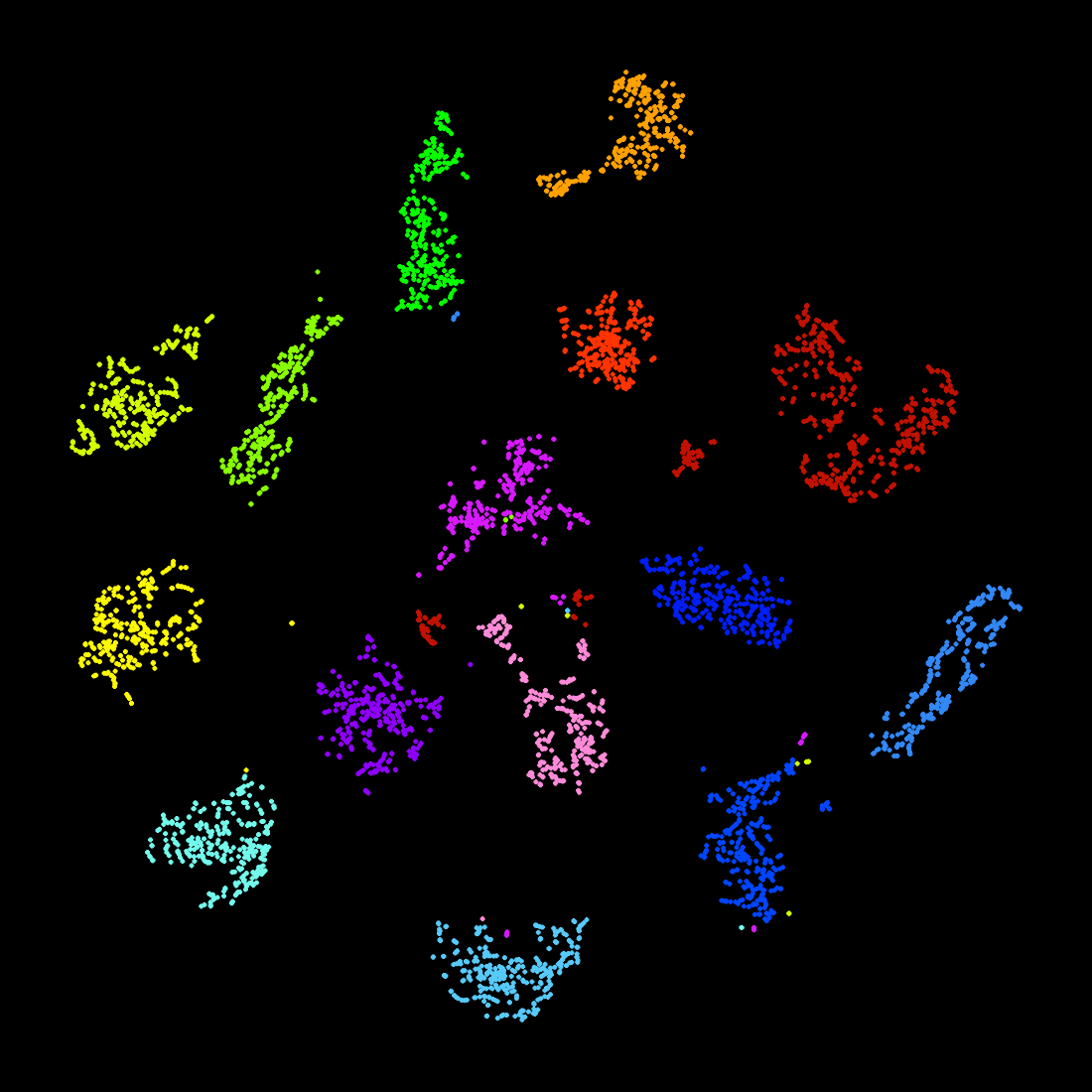}
		\includegraphics[scale=0.185]{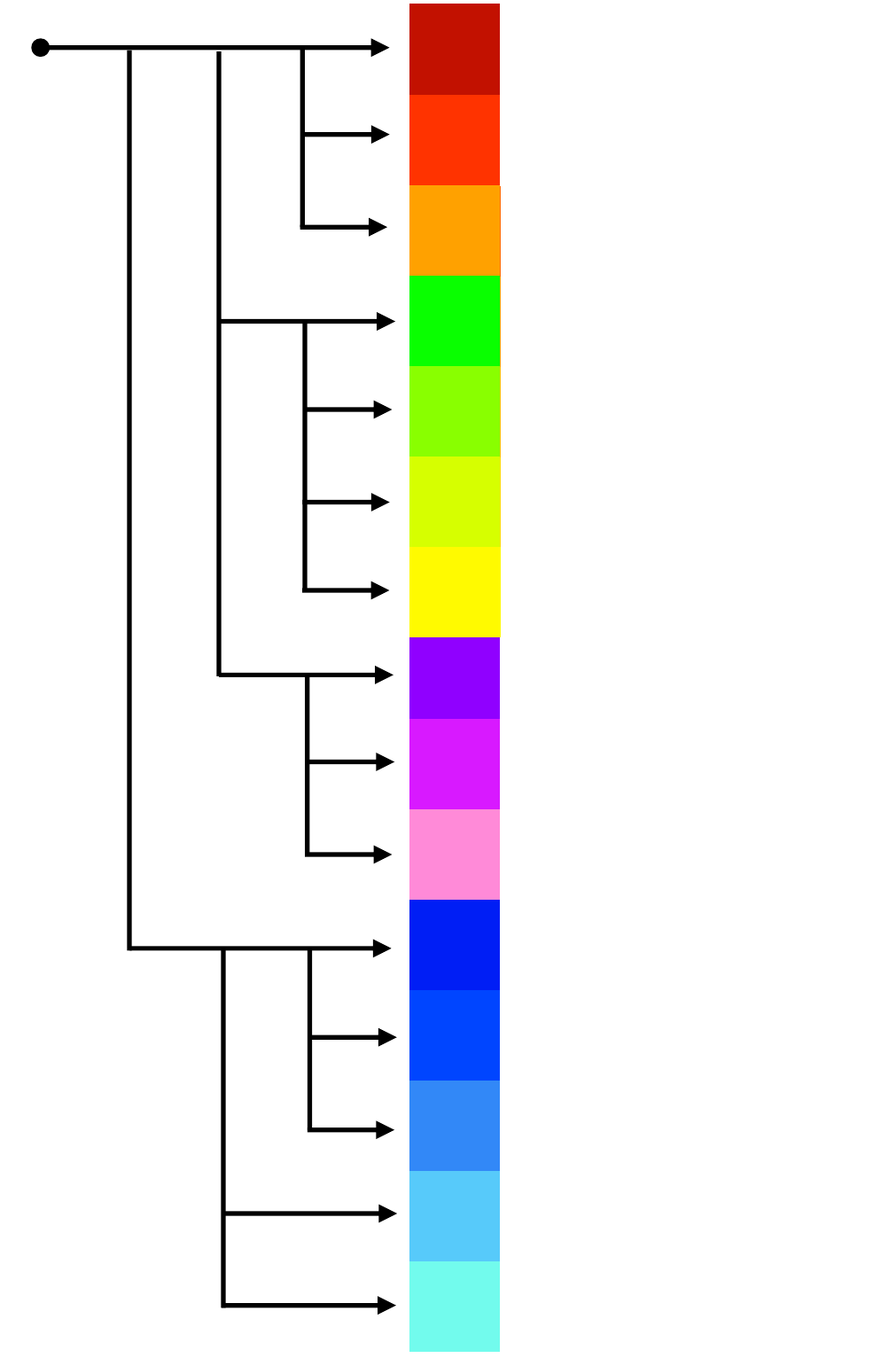}
		\put(-22.7,77.5){\tiny \textbf{bicyclist}}
		\put(-22.7,72.0){\tiny \textbf{motorcyclist}}
		\put(-22.7,66.5){\tiny \textbf{other-rider}}
		\put(-22.7,61.0){\tiny \textbf{bus}}
		\put(-22.7,55.5){\tiny \textbf{car}}
		\put(-22.7,50.0){\tiny \textbf{caravan}}
		\put(-22.7,44.5){\tiny \textbf{on-rails}}
		\put(-22.7,39.2){\tiny \textbf{sign-store}}
		\put(-22.7,34.0){\tiny \textbf{sign-advertise.}}
		\put(-22.7,28.6){\tiny \textbf{sign-informa.}}
		\put(-22.7,23.2){\tiny \textbf{tunnel}}
		\put(-22.7,17.8){\tiny \textbf{building}}
		\put(-22.7,12.3){\tiny \textbf{bridge}}
		\put(-22.7,6.8){\tiny \textbf{lake}}
		\put(-22.7,1.5){\tiny \textbf{road}}
		\end{tabular}
	\end{center}
	\vspace{-18pt}
	\captionsetup{font=small}
	\caption{\small\textbf{Visualization of the hierarchical embedding space} $f_{\text{ENC}}$ learned on Mapillary Vistas 2.0~\cite{neuhold2017mapillary} (\S\ref{sec:hrl}). The different colors correspond to different categories. It can be seen that, with $\mathcal{L}^{\text{TT}}$, $f_{\text{ENC}}$ (middle) nicely embraces the hierarchical semantic structures (right), in comparison with the one without $\mathcal{L}^{\text{TT}}$ (left).}
	\vspace{-12pt}
	\label{fig:embedding}
\end{figure}

Eq.~\ref{eq:TTL} encourages $f_{\text{ENC}}$ as a hierarchically-structured embedding space (Fig.~\ref{fig:embedding}): pixels with similar semantics (\ie,$_{\!}$ nearby in $\mathcal{T}$) are pushed closer than those with dissimilar semantics (\ie,$_{\!}$ faraway in $\mathcal{T}$), guided by the hierarchy-induced margin $m$. Related experiments are given in~\S\ref{sec:ablationstudy}.

\vspace{-2pt}
\subsection{Implementation Detail}\label{sec:detail}
\vspace{-2pt}

\noindent\textbf{Network Architecture.} \textsc{Hssn} is a general HSS framework; it is readily applied to any hierarchy-agnostic segmentation$_{\!}$ models. \textbf{i)} The \textit{segmentation encoder} $f_{\text{ENC}}$ (\S\ref{sec:hsss}) maps~each input image $I$ into a dense feature $\bm{I}\!\in\!\mathbb{R}^{H\!\times\!W\!\times\!C}$, and can be implemented as any backbone networks. In \S\ref{sec:4}, we experiment with two CNN-based (\ie, ResNet-101$_{\!}$~\cite{he2016deep} and HRNetV2-W48$_{\!}$~\cite{wang2020deep})  and a Transformer-based (\ie, Swin-Transformer~\cite{liu2021swin}) backbones. \textbf{ii)} The \textit{segmentation head} $f_{\text{SEG}}$  (\S\ref{sec:hsss}) projects $\bm{I}$ into a structured score map $\bm{S}\!\in\!\mathbb{R}^{H\!\times\!W\!\times\!|\mathcal{V}|}$ for all the classes in $\mathcal{V}$. Segmentation heads~used in recent segmentation models (\ie, DeepLabV3+$_{\!}$~\cite{chen2018encoder}, OCRNet$_{\!}$~\cite{yuan2020object}, MaskFormer$_{\!}$~\cite{cheng2021maskformer}) are used and modified.

\noindent\textbf{Training Objective.} \textsc{Hssn} is end-to-end trained by minimizing the combinatorial loss of our \textit{focal tree-min} loss ($\mathcal{L}^{\text{FTM}}$ in Eq.~\ref{eq:FocalTML}) and \textit{tree-triplet} loss  ($\mathcal{L}^{\text{TT}}$ in Eq.~\ref{eq:TTL}): $\mathcal{L}^{\text{FTM}} + \beta\mathcal{L}^{\text{TT}}$, where the coefficient $\beta\!\in\![0,0.5]$ is scheduled following a cosine annealing policy\!~\cite{loshchilov2016sgdr}. The focusing parameter $\gamma$ in $\mathcal{L}^{\text{FTM}\!}$ is set as 2. Furthermore, following the common practice in metric learning, a \textit{projection function} $f_{\text{PROJ}}$ is used in $\mathcal{L}^{\text{TT}}$. It maps each pixel embedding $\bm{i}$ into a 256-$d$ vector. $f_{\text{PROJ}}$ consists of two $1\!\times\!1$ convolutional layers and one ReLU between them, and is discarded after training, causing no extra computational cost in deployment.

\noindent\textbf{Inference.$_{\!}$} For$_{\!}$ each$_{\!}$ pixel,$_{\!}$ the$_{\!}$ label$_{\!}$ assignment$_{\!}$ follows$_{\!}$ Eq.$_{\!}$~\ref{eq:infer}.$_{\!\!\!}$

\begin{figure*}[t]
	\begin{center}
		\includegraphics[width=\linewidth]{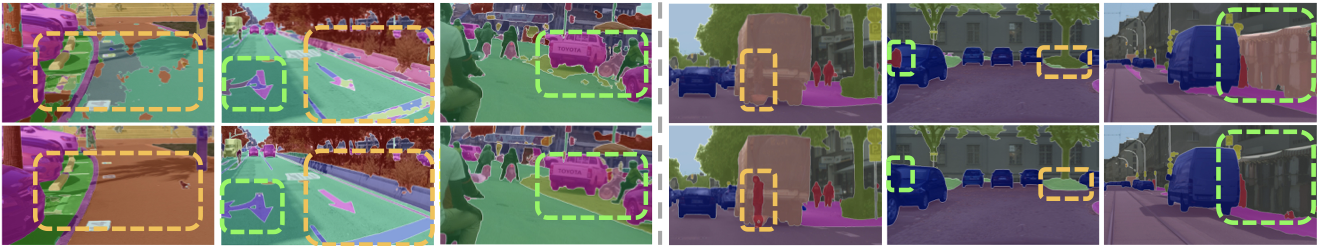}
	\end{center}
	\vspace{-18pt}
	\captionsetup{font=small}
	\caption{\small$_{\!}$\textbf{Visual$_{\!}$ results}$_{\!}$ (\S\ref{sec:vis})$_{\!}$ on$_{\!}$ Mapillary$_{\!}$ Vistas$_{\!}$ $2.0$\!~\cite{neuhold2017mapillary}$_{\!}$ \texttt{val}$_{\!}$ (left) and$_{\!}$ Cityscapes\!~\cite{cordts2016cityscapes}$_{\!}$ \texttt{val}$_{\!}$ (right).$_{\!}$ Top:$_{\!}$ MaskFormer,$_{\!}$ Bottom:$_{\!}$ \textsc{Hssn}.}
	\label{fig:sem}
	\vspace{-12pt}
\end{figure*}

\section{Experiment} \label{sec:4}

\subsection{Experimental Setup}

\noindent\textbf{Datasets.} We conduct experiments on two popular urban street scene
parsing datasets~\cite{neuhold2017mapillary,cordts2016cityscapes} and two human body  parsing datasets~\cite{xia2017joint,liang2018look}. The corresponding class hierarchies are either the officially provided  ones~\cite{neuhold2017mapillary,cordts2016cityscapes} or generated by following the conventions~\cite{xia2017joint,liang2018look}.
\begin{itemize}[leftmargin=*]
	\setlength{\itemsep}{0pt}
	\setlength{\parsep}{-2pt}
	\setlength{\parskip}{-0pt}
	\setlength{\leftmargin}{-10pt}
	\vspace{-6pt}
	
	\item \textbf{Mapillary$_{\!}$ Vistas$_{\!}$ 2.0}~\cite{neuhold2017mapillary} is an urban egocentric street-view dataset with high-resolution images. It contains $18,\!000$, $2,\!000$ and $5,\!000$ images for \texttt{train}, \texttt{val} and \texttt{test}, respectively. It provides annotations  for $144$ semantic concepts, which are organized in a  three-level hierarchy, covering $4/16/124$  concepts, respectively.

	\item \textbf{Cityscapes}~\cite{cordts2016cityscapes} contains $5,\!000$ elaborately annotated urban scene images, which are split into $2,\!975/500/1,\!524$ for \texttt{train}/\texttt{val}/\texttt{test}. It is associated with $19$ fine-grained  concepts, which are  grouped into $6$ super-classes.

	\item \textbf{PASCAL-Person-Part}~\cite{xia2017joint} has $1,\!716$ and $1,\!817$ images for \texttt{train} and \texttt{test}, with precise annotations for $6$ human parts. Following~\cite{wang2019learning,wang2020hierarchical}, we group $20$ fine-grained parts (\eg, \texttt{head}, \texttt{left-arm}) into two superclasses \texttt{upper-body} and \texttt{lower-body}, which are further combined ito \texttt{full-body}.
	
	\item \textbf{LIP}~\cite{liang2018look} includes $50,\!462$  single-person images gathered from real-world scenarios, with $30,\!462/10,\!000/10,\!000$ for $\texttt{train}/\texttt{val}/ \texttt{test}$ splits. The hierarchy is similar to the one in PASCAL-Person-Part, but the leaf layer has $19$ fine-grained semantic parts.
	\vspace{-4pt}
\end{itemize}

\noindent\textbf{Training.} For fair comparison, we follow~\cite{zhang2020blended,wang2019learning,zhao2017pyramid,chen2018encoder} to set the training hyper-parameters. Specifically, for CNN-based models, we use SGD as the optimizer with base learning rate 1e-2, momentum $0.9$ and weight decay 1e-4. For Transformer-based models,  we use AdamW~\cite{loshchilov2017decoupled}  with base  learning rate 6e-5 and weight decay $0.01$. The learning rate is scheduled by the polynomial annealing policy \cite{chen2017rethinking}. All backbones are initialized using the weights pre-trained on ImageNet-$1$K~\cite{deng2009imagenet}, while the remaining layers are randomly initialized. During training, we use standard data augmentation techniques, \ie, horizontal flipping and random scaling with a ratio between $0.5$ and $2.0$. We train $240$K and  $80$K iterations for Mapillary Vistas $2.0$ and Cityscapes, with batch size $8$ and crop size $512\!\times\!1024$. For  PASCAL-Person-Part and LIP, we use batch size $16$ and crop size $480\!\times\!480$, and train models  for $80$K and $160$K  iterations, respectively.

\noindent\textbf{Testing.} The inference follows  Eq.$_{\!}$~\ref{eq:infer}. As in \cite{huang2019ccnet,cheng2021maskformer,yuan2020object,ji2020learning,wang2020hierarchical,wang2019learning}, we report the segmentation scores at multiple scales ($\{0.5,0.75,1.0,1.25,1.5,1.75\}_{\!}$) with horizontal flipping.

\noindent\textbf{Evaluation Metric.} The mean {intersection-over-union} (mIoU) is adopted for evaluation. Particularly, we report the average score, \ie, mIoU$^l$, for classes in each hierarchy level $l$ independently.
For reference, we also report the scores of each level for hierarchy-agnostic methods. The results of each non-leaf layer are obtained by merging the segmentation predictions of its subclasses together.

\newcommand{\hssn}[2]{\setlength\tabcolsep{4pt}\renewcommand\arraystretch{1.05}
	\begin{tabular}{x{20}|y{20}}
		{#1} & {#2}
	\end{tabular}
}

\begin{table}
	\centering
	\small
	\resizebox{0.99\columnwidth}{!}{
		\setlength\tabcolsep{4pt}
		\renewcommand\arraystretch{1.05}
		\begin{tabular}{z{65}y{20}|c||ccc}
			\hline\thickhline
			\rowcolor{mygray}
			\multicolumn{2}{c|}{Method} & {Backbone} & $\text{mIoU}^3$$\uparrow$& $\text{mIoU}^2$$\uparrow$& $\text{mIoU}^1$$\uparrow$ \\ \hline\hline
			
			DeepLabV3+~\cite{chen2018encoder}\!\!&\!\!\pub{ECCV18}
			& ResNet-101  & 81.86 & 68.17 & 37.43 \\
			Seamless~\cite{porzi2019seamless}\!\!&\!\!\pub{CVPR19}
			& ResNet-101 & - & - &  38.17  \\
			OCRNet~\cite{yuan2020object}\!\!&\!\!\pub{ECCV20}
			&  HRNet-W48 & 83.19& 69.32& 38.26  \\
			HMSANet~\cite{wang2020hierarchical}\!\!&\!\!\pub{ArXiv19}
			& HRNet-W48 & 84.63& 70.71 & 39.53 \\
			MaskFormer~\cite{cheng2021maskformer}\!\!&\!\!\pub{NeurIPS21}
			& ResNet-101 &84.56 & 70.82& 39.60 \\
			MaskFormer~\cite{cheng2021maskformer}\!\!&\!\!\pub{NeurIPS21}
			& Swin-Small & 87.93 & 73.88 & 42.16 \\
			\hline
			
			\hssn{}{DeepLabV3+}  && ResNet-101 & {85.27}&  {71.40} & {40.16} \\
			\hssn{\textbf{\textsc{Hssn}}}{OCRNet} && HRNet-W48 & {86.46}& {72.34} & {41.13}   \\
			\hssn{}{MaskFormer} && Swin-Small & \textbf{90.02}& \textbf{75.81}& \textbf{43.97}   \\
			\hline
		\end{tabular}
	}
	\captionsetup{font=small}
	\caption{\small\textbf{Hierarchical semantic segmentation results}$_{\!}$ (\S\ref{sec:qr}) on the  \texttt{val} set of Mapillary Vistas $2.0$~\cite{neuhold2017mapillary}.}
	\label{table:mapillary}
	\vspace{-8pt}
\end{table}

\begin{table}
	\centering
	\small
	\resizebox{0.9\columnwidth}{!}{
		\setlength\tabcolsep{4pt}
		\renewcommand\arraystretch{1.05}
		\begin{tabular}{z{65}y{20}|c||cc}
			\hline\thickhline
			\rowcolor{mygray}
			\multicolumn{2}{c|}{Method} & {Backbone} & $\text{mIoU}^2$$\uparrow$& $\text{mIoU}^1$$\uparrow$ \\ \hline\hline
			DeepLabV2~\cite{chen2017deeplab}\!\!&\!\!\pub{CVPR17} & ResNet-101 & - & 70.22  \\
			PSPNet~\cite{zhao2017pyramid}\!\!&\!\!\pub{CVPR17} & ResNet-101 & - & 80.91 \\
			PSANet~\cite{zhao2018psanet}\!\!&\!\!\pub{ECCV18} & ResNet-101 & - & 80.96 \\
			PAN~\cite{li2018pyramid}\!\!&\!\!\pub{ArXiv18}& ResNet-101 & -& 81.12 \\
			{DeepLabV3}+~\cite{chen2018encoder}\!\!&\!\!\pub{ECCV18} & ResNet-101 &  92.16 & 82.08  \\
			DANet~\cite{fu2019dual}\!\!&\!\!\pub{CVPR19}& ResNet-101 & - &  81.52 \\
			Acfnet~\cite{zhang2019acfnet}\!\!&\!\!\pub{ICCV19} & ResNet-101 & - & 81.60 \\
			CCNet~\cite{huang2019ccnet}\!\!&\!\!\pub{ICCV19} & ResNet-101 & - &  81.08 \\
			HANet~\cite{choi2020cars}\!\!&\!\!\pub{CVPR20} & ResNet-101 & - &  81.82 \\
			HRNet~\cite{wang2020deep}\!\!&\!\!\pub{TPAMI20}&  HRNet-W48 & 92.12&  81.96 \\
			OCRNet~\cite{yuan2020object}\!\!&\!\!\pub{ECCV20}&  HRNet-W48 & 92.57& 82.33 \\
			MaskFormer~\cite{cheng2021maskformer}\!\!&\!\!\pub{NeurIPS21} & Swin-Small & 92.96 & 82.57  \\ \hline
			
			\hssn{}{DeepLabV3+}&
			&  ResNet-101 & {93.31} &{83.02} \\
			
			\hssn{\textbf{\textsc{Hssn}}}{OCRNet}&
			& HRNet-W48 & {93.92} &{83.37} \\
			\hssn{}{MaskFormer}&
			& Swin-Small & \textbf{94.39} & \textbf{83.74}\\
			\hline
		\end{tabular}
	}
	\captionsetup{font=small}
	\caption{\small\textbf{Hierarchical$_{\!}$ semantic segmentation results}$_{\!}$ (\S\ref{sec:qr})  on the  \texttt{val} set of Cityscapes~\cite{cordts2016cityscapes}.}
	\label{table:cityscape}
	\vspace{-15pt}
\end{table}

\begin{table*}
	\centering
	\small
	\resizebox{0.99\textwidth}{!}{
		\setlength\tabcolsep{6pt}
		\renewcommand\arraystretch{1.05}
		\begin{tabular}{z{65}y{20}||cccccccccc|ccc}
			\hline\thickhline
			\rowcolor{mygray}
			\multicolumn{2}{c||}{Method} & Head &  Torso & U-Arm & L-Arm & U-Leg & L-Leg & U-Body & L-Body & F-Body & B.G. & $\text{mIoU}^3$$\uparrow$& $\text{mIoU}^2$$\uparrow$& $\text{mIoU}^1$$\uparrow$ \\ \hline\hline
			DeepLabV3+~\cite{chen2018encoder}\!\!\!&\!\!\!\pub{ECCV18}\! & 87.02 & 72.02 & 60.37 & 57.36 & 53.54 & 48.52 & 90.07 & 65.88 & 93.02 & 96.07 & 94.55 & 84.01 &  67.84 \\
			SPGNet~\cite{cheng2019spgnet}\!\!\!&\!\!\!\pub{ICCV19}\! & 87.67 & 71.41 & 61.69 & 60.35 & 52.62 & 48.80 & - & - & - & 95.98 & - & - & 68.36 \\
			PGN~\cite{gong2019graphonomy}\!\!\!&\!\!\!\pub{CVPR19}\! & 90.89 & 75.12 & 55.83 & 64.61 & 55.42 & 41.57 & - & - & -& 95.33 & - & - & 68.40 \\
			CNIF~\cite{wang2019learning}\!\!\!&\!\!\!\pub{ICCV19}\!& 88.02 & 72.91 & 64.31 & 63.52 & 55.61 & 54.96 & 91.82 & 66.56 & 94.33 & 96.02 & 95.18 & 84.80 & 70.76\\
			SemaTree~\cite{ji2020learning}\!\!\!&\!\!\!\pub{ECCV20}\! & 89.15 & 74.76 & 63.90 &63.95 & 57.53 & 54.62 & 92.36 & 67.13 & 95.11& 96.84 & 95.98 & 85.44 & 71.59 \\
			HHP~\cite{wang2020hierarchical}\!\!\!&\!\!\!\pub{CVPR20}\! & 89.73 & 75.22 & 66.87 & 66.21 & 58.69 & 58.17 & 93.44 & 68.02 & 96.77 & 96.94 & 96.86 & 86.13 & 73.12 \\
			BGNet~\cite{zhang2020blended}\!\!\!&\!\!\!\pub{ECCV20}\! & 90.18 & 77.44 & 68.93 & 67.15 & 60.79 & 59.27 & - & - & -& 97.12 & - & - & 74.42 \\
			PCNet~\cite{zhang2020part}\!\!\!&\!\!\!\pub{CVPR20}\!& 90.04 & 76.89 & 69.11 & 68.40 & 60.78 & 60.14 & - & - & -& 96.78 & - & - & 74.59 \\ \hline
			\hssn{\textbf{\textsc{Hssn}}}{DeepLabV3+}& & \textbf{90.19}  &  \textbf{78.72} & \textbf{70.67}& \textbf{69.71}& \textbf{61.15}& \textbf{60.44}& \textbf{95.86} & \textbf{71.56} & \textbf{98.20} & \textbf{97.18} & \textbf{97.69} & \textbf{88.20} & \textbf{75.44} \\
			\hline
		\end{tabular}
	}
	\captionsetup{font=small}
	\caption{\small\textbf{Hierarchical human parsing results} (\S\ref{sec:qr}) on PASCAL-Person-Part~\cite{xia2017joint} \texttt{test}. All models use ResNet-101 as the backbone.}
	\label{table:ppp}
	\vspace{-15pt}
\end{table*}

\begin{table}
	\centering
	\small
	\resizebox{\columnwidth}{!}{
		\setlength\tabcolsep{4pt}
		\renewcommand\arraystretch{1.05}
		\begin{tabular}{z{65}y{22}|c||ccc}
			\hline\thickhline
			\rowcolor{mygray}
			\multicolumn{2}{c|}{Method} & {Backbone} & $\text{mIoU}^3$$\uparrow$&$\text{mIoU}^2$$\uparrow$& $\text{mIoU}^1$$\uparrow$ \\ \hline\hline
			SegNet~\cite{badrinarayanan2017segnet}\!\!&\!\!\pub{TPAMI17}\! & ResNet-101& - & - & 18.17 \\
			FCN-8s~\cite{long2015fully}\!\!&\!\!\pub{CVPR15}\! & ResNet-101& - & - & 28.29 \\
			DeepLabV2~\cite{chen2017deeplab}\!\!&\!\!\pub{CVPR17}\!& ResNet-101& - & - & 41.64 \\
			Attention~\cite{chen2016attention}\!\!&\!\!\pub{CVPR16}\! & ResNet-101& - & - & 42.92 \\
			MMAN~\cite{luo2018macro}\!\!&\!\!\pub{ECCV18}\! & ResNet-101& - & - & 46.93  \\
			DeepLabV3+\!~\cite{chen2018encoder}\!\!&\!\!\pub{ECCV18}\! & ResNet-101 & 88.13 & 83.97 & 52.28 \\
			CE2P~\cite{ruan2019devil}\!\!&\!\!\pub{AAAI19}\! & ResNet-101& - & - & 53.10 \\
			BraidNet~\cite{liu2019braidnet}\!\!&\!\!\pub{ACMMM19}\! & ResNet-101& - & - &54.42 \\
			SemaTree~\cite{ji2020learning}\!\!&\!\!\pub{ECCV20}\! & ResNet-101& 90.78 & 87.12 & 54.73 \\
			BGNet~\cite{zhang2020blended}\!\!&\!\!\pub{ECCV20}\! & ResNet-101& - & - & 56.82 \\
			PCNet~\cite{zhang2020part}\!\!&\!\!\pub{CVPR20}\! & ResNet-101& - & - & 57.03 \\
			CNIF~\cite{wang2019learning}\!\!&\!\!\pub{ICCV19}\! & ResNet-101& 95.92 & 91.83 & 57.74 \\
			HRNet~\cite{wang2020deep}\!\!&\!\!\pub{TPAMI20}\! &  HRNet-W48 & 95.53&91.21 &  57.23 \\
			OCRNet~\cite{yuan2020object}\!\!&\!\!\pub{ECCV20}\! &  HRNet-W48 & 96.78& 92.56 &  58.47 \\
			HHP~\cite{wang2020hierarchical}\!\!&\!\!\pub{CVPR20}\!  & ResNet-101& 97.41 & 93.43 & 59.25 \\ \hline
			\hssn{\textbf{\textsc{Hssn}}}{DeepLabV3+}& & ResNet-101&  \textbf{98.86} & \textbf{94.75} & \textbf{60.37} \\
			\hline
		\end{tabular}
	}
	\captionsetup{font=small}
	\caption{\small$_{\!}$\textbf{Hierarchical$_{\!}$ human$_{\!}$ parsing$_{\!}$ results} (\S\ref{sec:qr}) on LIP \texttt{val}.$_{\!\!\!}$}
	\label{table:lip}
	\vspace{-16pt}
\end{table}

\subsection{Quantitative Results} \label{sec:qr}

\noindent\textbf{Mapillary Vistas 2.0~\cite{neuhold2017mapillary}.} Table \ref{table:mapillary} presents comparisons of our {\textsc{Hssn}} against several top-leading semantic segmentation models on Mapillary Vistas $2.0$ \texttt{val}. With the standard ResNet-101 as the backbone, {\textsc{Hssn}} outperforms the famous DeepLabV3+~\cite{chen2018encoder} by solid margins across all three levels (\textbf{2.69\%}/\textbf{3.21\%}/\textbf{3.40\%}). Consistent gains are also observed for a more recent segmentation model (\ie, MaskFormer \cite{cheng2021maskformer}), which relies on a heavy Transformer-based decoder. In addition, our {\textsc{Hssn}} further improves the performance when using more advanced CNN-based  (\ie, HRNetV2-W48) or  Transformer-based (\ie, Swin-Small) backbones. Concretely, it outperforms OCRNet~\cite{yuan2020object} by \textbf{2.87\%}/\textbf{3.02\%}/\textbf{3.27\%} and MaskFormer \cite{cheng2021maskformer} by \textbf{1.81\%}/ \textbf{1.93\%}/\textbf{2.09\%} across the three levels. {\textsc{Hssn}}, with Swin-Small as the backbone, establishes a new state-of-the-art. These results clearly demonstrate the efficacy of our hierarchical semantic segmentation framework.

\begin{figure*}[t]
\vspace{-5pt}
	\begin{center}
		\includegraphics[width=\linewidth]{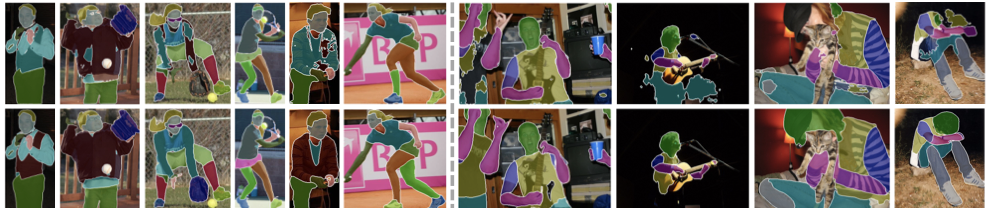}
	\end{center}
	\vspace{-18pt}
	\captionsetup{font=small}
	\caption{\small\textbf{Visual results} (\S\ref{sec:vis}) on LIP~\cite{liang2018look} \texttt{val} (left) and PASCAL-Person-Part~\cite{xia2017joint} \texttt{test} (right). Top: DeepLabV3+, Bottom: \textsc{Hssn}.}
	\label{fig:hp}
	\vspace{-16pt}
\end{figure*}

\noindent\textbf{Cityscapes~\cite{cordts2016cityscapes}.} Table \ref{table:cityscape} compares our {\textsc{Hssn}} with several competitive models on Cityscapes \texttt{val}. Despite that the dataset has relatively simple semantic hierarchy and has been comprehensively benchmarked, our model still leads to appealing improvements. In particular, {\textsc{Hssn}} outperforms the top-leading MaskFormer \cite{cheng2021maskformer} by \textbf{1.17\%}/\textbf{1.43\%} in terms of mIoU$^1$ and mIoU$^2$ when using Swin-Small as the backbone. Similar gains are obtained when applying CNN-based backbones (\ie, ResNet-101 and HRNet-W48).

\noindent\textbf{PASCAL-Person-Part~\cite{xia2017joint}.}
Table \ref{table:ppp} lists the detailed results on PASCAL-Person-Part \texttt{test}. Note that all the models use ResNet-101 as the backbone. As seen, our {\textsc{Hssn}}  achieves the best performance for all human parts and hierarchical levels.
 Remarkably, \textsc{Hssn} outperforms all existing hierarchical human parsers (\ie, HHP~\cite{wang2020hierarchical}, SemaTree~\cite{ji2020learning} and CNIF~\cite{wang2019learning}) by significant margins. Results on this dataset are particularly impressive since it includes a very small number (\ie, $1,\!713$) of  training samples.

\noindent\textbf{LIP~\cite{liang2018look}.} In Table \ref{table:lip}, we compare {\textsc{Hssn}} with state-of-the-art human semantic parsing models on LIP \texttt{val}. As observed, our model provides a considerable performance gain against the leading hierarchy-aware human parser (\ie, HHP~\cite{wang2020hierarchical}) across all three levels ($\textbf{1.12\%}/\textbf{1.32\%}/\textbf{1.45\%}$). These results support our motivation of exploiting structured label constraints and structured representation learning rather than only focusing on structured feature fusion.

\subsection{Qualitative Results} \label{sec:vis}
Fig.~\ref{fig:sem} and Fig.~\ref{fig:hp} depict representative visual results on four datasets. As seen, \textsc{Hssn} yields more precise segmentation results in comparison with some top-performing methods (\ie, MaskFormer in Fig.~\ref{fig:sem} and DeepLabV3+ in Fig.~\ref{fig:hp}), and shows strong robustness to various challenging scenarios with occlusions, small objects and densely arranged targets, \etc. Moreover, as shown in the last column of Fig.~\ref{fig:sem}, MaskFormer makes a severe mistake that misclassifies a part of background structure  as \texttt{truck}. In contrast, benefiting from hierarchy-aware segmentation learning, \textsc{Hssn} naturally address the issue of mistake severity, \ie, distinguish significantly different concepts with larger margins.

\subsection{Diagnostic Experiment} \label{sec:ablationstudy}

To gain more insights into \textsc{Hssn}, we conduct a set of ablative studies on Mapillary Vistas 2.0~\cite{neuhold2017mapillary} and Pascal-Person-Part~\cite{xia2017joint}, with ResNet-101 as the backbone.

\noindent\textbf{Key Component Analysis.}
First, we investigate the essential designs in \textsc{Hssn}, \ie,  hierarchical segmentation learning (\S\ref{sec:hsl}) with $\mathcal{L}^{\text{FTM}\!}$ (\textit{cf.}$_{\!}$~Eq.~\ref{eq:FocalTML}) and hierarchical representation  learning (\S\ref{sec:hrl}) with $\mathcal{L}^{\text{TT}\!}$ (\textit{cf.}$_{\!}$~Eq.~\ref{eq:TTL}). The results are summarized in Table \ref{table:ablation3}.
The first row refers to a hierarchy-agnostic baseline that only concerns the leaf nodes and  is trained using the categorical cross-entropy loss $\mathcal{L}^{\text{CCE}}$ (\textit{cf.}$_{\!}$~Eq.~\ref{eq:CCE}). Three crucial conclusions can be drawn.  \textbf{First}, our $\mathcal{L}^{\text{FTM}\!}$ leads to significant performance improvements against the baseline across all the metrics on both datasets.  This evidences that our hierarchical segmentation learning strategy is able to produce hierarchy-coherent predictions. \textbf{Second},  we also observe compelling gains by incorporating $\mathcal{L}^{\text{TT}}$ into the baseline. This proves  the importance of hierarchical representation  learning. \textbf{Third}, our full model achieves the best performance by combining our $\mathcal{L}^{\text{FTM}}$  and $\mathcal{L}^{\text{TT}}$ together, confirming the necessity of joint hierarchical segmentation and embedding learning.

\noindent\textbf{Focal Tree-Min Loss.} We next examine the design of our focal tree-min loss $\mathcal{L}^{\text{FTM}}$ (\textit{cf.}$_{\!}$~Eq.~\ref{eq:FocalTML}). As shown in Table~\ref{table:ablation1}, we compare $\mathcal{L}^{\text{FTM}}$ with four different losses, \ie, categorical cross-entropy loss $\mathcal{L}^{\text{CCE}}$ (\textit{cf.}$_{\!}$~Eq.~\ref{eq:CCE}), binary cross-entropy loss $\mathcal{L}^{\text{BCE}}$ (\textit{cf.}$_{\!}$~Eq.~\ref{eq:sigmoidbce}), focal loss~\cite{lin2017focal}, and our tree-min loss $\mathcal{L}^{\text{TM}}$ (\textit{cf.}$_{\!}$~Eq.~\ref{eq:TML}). We can find that our $\mathcal{L}^{\text{TM}}$  generates impressive results, and $\mathcal{L}^{\text{FTM}}$ is even better than $\mathcal{L}^{\text{TM}}$. Then, in Table~\ref{table:gamma}, we analyze the impact of the focusing parameter $\gamma$ in $\mathcal{L}^{\text{FTM}}$. As seen, the performance progressively improves as $\gamma$ is increased, and the gain becomes marginal when $\gamma\!=\!2$. Hence, we choose $\gamma\!=\!2$ by default.

\begin{table}
	\centering
	\small
	\resizebox{\columnwidth}{!}{
		\setlength\tabcolsep{4pt}
		\renewcommand\arraystretch{1.1}
		\begin{tabular}{cc|c>{\columncolor[gray]{0.95}}c>{\columncolor[gray]{0.90}}c||c>{\columncolor[gray]{0.95}}c>{\columncolor[gray]{0.90}}c}
        \hline\thickhline
		     \rowcolor{mygray}
		     $\mathcal{L}^{\text{FTM}}$&$\mathcal{L}^{\text{TT}}$  & \multicolumn{3}{c||}{Mapillary Vistas 2.0} & \multicolumn{3}{c}{Pascal-Person-Part} \\ \cline{3-8}
            \rowcolor{mygray}
            Eq.~\ref{eq:FocalTML} &Eq.~\ref{eq:TTL} & $\text{mIoU}^3$$\uparrow$& $\text{mIoU}^2$$\uparrow$& $\text{mIoU}^1$$\uparrow$& $\text{mIoU}^3$$\uparrow$& $\text{mIoU}^2$$\uparrow$& $\text{mIoU}^1$$\uparrow$ \\ \hline\hline
			 &  & 81.86 & 68.17 & 37.43 & 93.58 & 83.04 & 67.84 \\ \hline
            \cmark &  &84.17& 69.62& 39.17& 96.33 & 86.72 & 72.89\\
             & \cmark & 83.06 & 68.61 & 38.29 & 95.92 & 86.03 & 72.27\\
			\cmark & \cmark  & \textbf{85.27} & \textbf{71.40} &\textbf{40.16}& \textbf{97.69} & \textbf{88.20} & \textbf{75.44}\\
			\hline
		\end{tabular}
	}
	\captionsetup{font=small}
	\caption{\small \textbf{Analysis of essential components} on Mapillary Vistas $2.0$~\cite{neuhold2017mapillary} \texttt{val} and PASCAL-Person-Part~\cite{xia2017joint} \texttt{test} (\S\ref{sec:ablationstudy}).}
	\label{table:ablation3}
	\vspace{-8pt}
\end{table}

\begin{table}
	\centering
	\small
	\resizebox{\columnwidth}{!}{
		\setlength\tabcolsep{4pt}
		\renewcommand\arraystretch{1.05}
		\begin{tabular}{c||c>{\columncolor[gray]{0.95}}c>{\columncolor[gray]{0.90}}c|c>{\columncolor[gray]{0.95}}c>{\columncolor[gray]{0.90}}c}
			\hline\thickhline
		     \rowcolor{mygray}
             & \multicolumn{3}{c|}{Mapillary Vistas 2.0} & \multicolumn{3}{c}{Pascal-Person-Part} \\ \cline{2-7}
            \rowcolor{mygray}
             \multirow{-2}{*}{Loss} &$\text{mIoU}^3$$\uparrow$& $\text{mIoU}^2$$\uparrow$& $\text{mIoU}^1$$\uparrow$ & $\text{mIoU}^3$$\uparrow$& $\text{mIoU}^2$$\uparrow$& $\text{mIoU}^1$$\uparrow$ \\ \hline\hline
             CCE & 81.86 & 68.17 & 37.43 & 93.58 & 83.04 & 67.84 \\
             BCE & 81.56 & 67.61 & 37.26 & 93.12 & 82.55 & 67.38\\
             Focal  &82.63& 68.48& 38.09  & 94.07 & 83.66 & 68.42\\ \hline
             TM &83.48 &69.13 & 38.69 & 95.32 & 85.99 & 72.17\\
             FTM &84.17& 69.62& 39.17& 96.33 & 86.72 & 72.89 \\ \hline
             Full & \textbf{85.27} & \textbf{71.40} &\textbf{40.16}&\textbf{97.69} & \textbf{88.20} & \textbf{75.44}\\
			\hline
		\end{tabular}
	}
	\captionsetup{font=small}
	\caption{\small \textbf{Analysis of focal tree-min loss $\mathcal{L}^{\text{FTM}}$} on Mapillary Vistas $2.0$~\cite{neuhold2017mapillary} \texttt{val} and PASCAL-Person-Part~\cite{xia2017joint} \texttt{test} (\S\ref{sec:ablationstudy}).}
	\label{table:ablation1}
	\vspace{-12pt}
\end{table}

\noindent\textbf{Tree-Triplet Loss.} We further investigate the design of our tree-triplet loss $\mathcal{L}^{\text{TT}\!}$ (\textit{cf.}$_{\!}$~Eq.~\ref{eq:TTL}). In Table \ref{table:ablation4},  ``Vanilla'' refers to the vanilla triplet loss with a constant margin \cite{schroff2015facenet}. By constructing hierarchy-aware triplet samples, our tree-triplet loss $\mathcal{L}^{\text{TT}}$ (also with a constant margin) outperforms ``Vanilla''. The gains become larger when further applying the hierarchy-induced margin constraint. These results confirm the designs of our tree-triplet loss. Finally, we assess the impact of the distance measurement $\langle\cdot,\cdot\rangle$ used in $\mathcal{L}^{\text{TT}\!}$. We study Cosine and Euclidean distances. Table \ref{table:ablation2} shows that Cosine distance performs much better than Euclidean distance, corroborating relevant observations in \cite{nickel2017poincare,ganea2018hyperbolic,sala2018representation}.

\begin{table}
	\centering
	\small
	\resizebox{\columnwidth}{!}{
		\setlength\tabcolsep{5pt}
		\renewcommand\arraystretch{1.03}
		\begin{tabular}{c||c>{\columncolor[gray]{0.95}}c>{\columncolor[gray]{0.90}}c|c>{\columncolor[gray]{0.95}}c>{\columncolor[gray]{0.90}}c}
			\hline\thickhline
			\rowcolor{mygray}
			$\gamma$& \multicolumn{3}{c|}{Mapillary Vistas 2.0} & \multicolumn{3}{c}{Pascal-Person-Part} \\ \cline{2-7}
			\rowcolor{mygray}
			Eq.~\ref{eq:FocalTML} & $\text{mIoU}^3$$\uparrow$& $\text{mIoU}^2$$\uparrow$& $\text{mIoU}^1$$\uparrow$& $\text{mIoU}^3$$\uparrow$& $\text{mIoU}^2$$\uparrow$& $\text{mIoU}^1$$\uparrow$ \\ \hline\hline
			$0$ &84.47 &70.24& 39.52 & 96.90 &  87.56 & 74.84 \\
			$0.2$ & 84.53&70.38& 39.62 & 97.17 & 87.71 & 74.91\\
			$0.5$ & 84.85&70.61& 39.72 & 97.23 & 87.68 & 74.94\\
			$1.0$ & 85.11 &70.95& 39.94 & 97.44 & 87.97 & 75.20\\
			$2.0$ & \textbf{85.27} & \textbf{71.40} &\textbf{40.16}& \textbf{97.69} & \textbf{88.20} & \textbf{75.44}\\
			$5.0$ & 84.92& 70.07 & 39.40 & 96.84 & 87.25 & 74.65\\
			\hline
		\end{tabular}
	}
	\captionsetup{font=small}
	\caption{\small \textbf{Analysis of $\gamma$} for $\mathcal{L}^{\text{FTM}}$ (Eq.~\ref{eq:FocalTML}) on Mapillary Vistas $2.0$~\cite{neuhold2017mapillary} \texttt{val} and PASCAL-Person-Part~\cite{xia2017joint} \texttt{test} (\S\ref{sec:ablationstudy}).}
	\label{table:gamma}
	\vspace{-8pt}
\end{table}

\begin{table}
	\centering
	\small
	\resizebox{\columnwidth}{!}{
		\setlength\tabcolsep{2pt}
		\renewcommand\arraystretch{1.05}
		\begin{tabular}{c|c||c>{\columncolor[gray]{0.95}}c>{\columncolor[gray]{0.90}}c|c>{\columncolor[gray]{0.95}}c>{\columncolor[gray]{0.90}}c}
			\hline\thickhline
			\rowcolor{mygray}
			Triplet&Margin & \multicolumn{3}{c|}{Mapillary Vistas 2.0} & \multicolumn{3}{c}{Pascal-Person-Part} \\ \cline{3-8}
			\rowcolor{mygray}
			Loss & $m$ & $\text{mIoU}^3$$\uparrow$& $\text{mIoU}^2$$\uparrow$& $\text{mIoU}^1$$\uparrow$& $\text{mIoU}^3$$\uparrow$& $\text{mIoU}^2$$\uparrow$& $\text{mIoU}^1$$\uparrow$ \\ \hline\hline
			Vanilla
			& Constant & 84.25&70.13& 39.41 & 96.58 & 87.03&74.10 \\ \hline
			
			$\mathcal{L}^{\text{TT}}$
			& Constant & 84.66&70.42& 39.67 & 97.30 & 87.86 & 74.83\\
			
			$\mathcal{L}^{\text{TT}}$
			 & Hierarchy& \textbf{85.27} & \textbf{71.40} &\textbf{40.16}& \textbf{97.69} & \textbf{88.20} & \textbf{75.44}\\
			\hline
		\end{tabular}
	}
	\captionsetup{font=small}
	\caption{\small \textbf{Analysis of different variants of $\mathcal{L}^{\text{TT}}$} on Mapillary Vistas $2.0$~\cite{neuhold2017mapillary} \texttt{val} and PASCAL-Person-Part~\cite{xia2017joint} \texttt{test} (\S\ref{sec:ablationstudy}).}
	\label{table:ablation4}
	\vspace{-8pt}
\end{table}

\begin{table}
	\centering
	\small
	\resizebox{\columnwidth}{!}{
		\setlength\tabcolsep{2pt}
		\renewcommand\arraystretch{1.05}
		\begin{tabular}{c||c>{\columncolor[gray]{0.95}}c>{\columncolor[gray]{0.90}}c|c>{\columncolor[gray]{0.95}}c>{\columncolor[gray]{0.90}}c}
			\hline\thickhline
			\rowcolor{mygray}
			Distance&  \multicolumn{3}{c|}{Mapillary Vistas 2.0} & \multicolumn{3}{c}{Pascal-Person-Part}\\ \cline{2-7}
			\rowcolor{mygray}
			{Measurement} & $\text{mIoU}^3$$\uparrow$& $ \text{mIoU}^2$$\uparrow$& $\text{mIoU}^1$$\uparrow$ & $\text{mIoU}^3$$\uparrow$& $\text{mIoU}^2$$\uparrow$& $\text{mIoU}^1$$\uparrow$ \\ \hline\hline
			Euclidean  & 84.23& 70.02& 39.33 &96.28 &86.73& 73.88 \\
			Cosine &  \textbf{85.27} & \textbf{71.40} &\textbf{40.16}& \textbf{97.69} & \textbf{88.20} & \textbf{75.44}\\
			\hline
		\end{tabular}
	}
	\captionsetup{font=small}
	\caption{\small \textbf{Analysis of distance measure} for  $\mathcal{L}^{\text{TT}}$ on Mapillary Vistas $2.0$~\cite{neuhold2017mapillary} \texttt{val} and PASCAL-Person-Part~\cite{xia2017joint} \texttt{test} (\S\ref{sec:ablationstudy}).}
	\label{table:ablation2}
	\vspace{-15pt}
\end{table}

\section{Conclusion}
In this paper, we presented \textsc{Hssn}, a structured solution for semantic segmentation.  \textsc{Hssn} is capable of exploiting taxonomic semantic relations for structured scene parsing, by only slightly changing existing hierarchy-agnostic segmentation networks. By exploiting hierarchy properties as optimization criteria, hierarchical violation in the segmentation predictions can be explicitly penalized. Through hierarchy-induced margin separation, more effective pixel representations can be generated. We experimentally show that \textsc{Hssn} outperforms many existing segmentation models on four famous datasets.
 We wish this work to pave the way for future research on hierarchical semantic segmentation.

{\small
\bibliographystyle{ieee_fullname}
\bibliography{egbib}
}

\clearpage
\appendix
\section{Detailed Hierarchical Architecture}
We use the official hierarchical structure provided in each dataset. Detailed semantic hierarchies are provided in Fig.$_{\!}$~\ref{fig:mapillary_h} for Mapillary Vistas 2.0~\cite{neuhold2017mapillary}, Fig.$_{\!}$~\ref{fig:cityscape_h} for Cityscapes~\cite{cordts2016cityscapes}, Fig.$_{\!}$~\ref{fig:ppp_h} for PASCAL-Person-Part~\cite{xia2017joint} and Fig.$_{\!}$~\ref{fig:lip_h} for LIP~\cite{liang2018look}. For Mapillary Vistas 2.0 and Cityscapes, we add a virtual root node (\ie, \texttt{All}) to represent the most general concept.

\begin{table}[b]
\renewcommand\thetable{C.1}
	\centering
	\small
    \subfloat[\label{a}]{
        \resizebox{0.65\columnwidth}{!}{
        		\begin{tabular}{c||cc}
        \hline\thickhline
            \rowcolor{mygray}
            {$m_{\varepsilon}$ in Eq.~\ref{eq:margin}} & $\text{mIoU}^2$$\uparrow$ & $\text{mIoU}^1$$\uparrow$ \\ \hline\hline
			0.05& 93.16 {\color{mywarning}{(\textbf{-0.13})}} & 82.81 {\color{mywarning}{(\textbf{-0.18})}}   \\
			\rowcolor{myorange}
			0.10 (default) & 93.29 & 82.99\\
			0.15& 93.19 {\color{mywarning}{(\textbf{-0.10})}}& 82.86 {\color{mywarning}{(\textbf{-0.13})}} \\
			\hline
		\end{tabular}
        }
        }
        \vspace{-10pt}
    \subfloat[\label{b}]{
       \resizebox{0.65\columnwidth}{!}{
        		\begin{tabular}{c||cc}
        \hline\thickhline
            \rowcolor{mygray}
            {0.5 in Eq.~\ref{eq:margin}} &$\text{mIoU}^2$$\uparrow$ & $\text{mIoU}^1$$\uparrow$ \\ \hline\hline
			0.45& 93.17 {\color{mywarning}{(\textbf{-0.12})}} & 82.84 {\color{mywarning}{(\textbf{-0.15})}} \\
			\rowcolor{myorange}
			0.50 (\textit{default}) & 93.29 & 82.99 \\
			0.55& 93.13 {\color{mywarning}{(\textbf{-0.16})}} & 82.80 {\color{mywarning}{(\textbf{-0.19})}} \\
			\hline
		\end{tabular}
        }
        }
    \vspace{-10pt}
    \subfloat[\label{c}]{
		\resizebox{0.95\columnwidth}{!}{
			\setlength\tabcolsep{12pt}
			\renewcommand\arraystretch{1.}
			\begin{tabular}{c|c||cc}
				\hline\thickhline
				\rowcolor{mygray}
				$\beta$ {Schedule} & {Value} & $\text{mIoU}^2$$\uparrow$& $\text{mIoU}^1$$\uparrow$ \\ \hline\hline
		        \rowcolor{myorange}
				Cosine (\textit{default}) & 0 $\rightarrow$ 0.5 & 93.29 & 82.99  \\
				Constant & 0.5 & 93.02 {\color{mywarning}{(\textbf{-0.27})}} & 82.68 {\color{mywarning}{(\textbf{-0.31})}} \\ \hline
			\end{tabular}
		}
	}
	\vspace{-5pt}
	\captionsetup{font=small}
	\caption{\small\textbf{More ablative experiments for hyper-parameters}\!~(\S\ref{sc:ablative}) on Cityscapes\!~\cite{cordts2016cityscapes} \texttt{val}.}
	\label{table:ablative}
    \vspace{-10pt}
\end{table}

\section{Additional Qualitative Result}
We provide additional visualization results on four datasets, including Mapillary Vistas 2.0~\cite{neuhold2017mapillary} \texttt{val} in Fig.$_{\!}$~\ref{fig:mapillary}, Cityscapes~\cite{cordts2016cityscapes} \texttt{val} in Fig.$_{\!}$~\ref{fig:cityscape}, PASCAL-Person-Part~\cite{xia2017joint} \texttt{val} in Fig.$_{\!}$~\ref{fig:ppp} and LIP~\cite{liang2018look} \texttt{val} in Fig.$_{\!}$~\ref{fig:lip}. The left column shows results from the baseline model while the right column is the predictions produced by \textsc{Hssn}. We see that \textsc{Hssn} yields consistently better visual effects than the baseline model.

\section{Additional Ablative Study} \label{sc:ablative}
We give extra ablative studies for the hyper-parameters emerged in our approach in Table \ref{table:ablative}. It can be seen that, for {$m_{\varepsilon}$ and {0.5} in Eq.~\ref{eq:margin}, there is minor impact to the perfromance. This indicates our method is robust to hyper-parameters. For the balance factor $\beta$ between $\mathcal{L}^\text{FTM}$ and $\mathcal{L}^\text{TT}$, scheduling it in a cosine annealing policy yields better performance. It is reasonable due to the poor recognition capability of the network at the initial stage of training.

\section{Discussion on Triplet Number} \label{sc:lim}
We further investigate the impact of the number of triplets in $\mathcal{L}^\text{TT}$ sampled druing training  on performance. It can be seen in Table \ref{table:trip} that the introduce of triplet loss imposes additional computation to the model and slows down the training speed. However, \textsc{Hssn} is able to reach very promising performance using a small number of triplets  (\eg, 200) on both datasets. Further increasing the number only brings minor improvements. Based on the results in Table~\ref{table:trip}, we set the number to 200 for all datasets.
This facilitates \textsc{Hssn} to perform triplet sampling at negligible cost and be rewarded with impressive performance boost.

\begin{table}
\renewcommand\thetable{D.1}
	\centering
	\small
	\resizebox{\columnwidth}{!}{
		\setlength\tabcolsep{3pt}
		\renewcommand\arraystretch{1.05}
		\begin{tabular}{c|c|cc||cc}
			\hline\thickhline
		     \rowcolor{mygray}
		     & & \multicolumn{2}{c||}{Mapillary Vistas 2.0} & \multicolumn{2}{c}{{PASCAL-Person-Part}} \\ \cline{3-6}
            \rowcolor{mygray}
            \multirow{-2}{*}{\#} & \multirow{-2}{*}{\# Triplets}  & mIoU$\uparrow$ & Time $\downarrow$ & mIoU $\uparrow$ & Time $\downarrow$\\ \hline\hline
			1 &  -- & 39.17 & 0.93 & 72.89 & 0.41 \\ \hline
			2 & 100 & 39.82 \color{mywarning}{(\textbf{-0.34})} & 1.06 \color{mygreen}{(\textbf{-0.12})} & 74.76 \color{mywarning}{(\textbf{-0.68})} & 0.52 \color{mygreen}{(\textbf{-0.20})} \\
			3 & 200  (\textit{default}) & \cellcolor{myorange}\textbf{40.16} & \cellcolor{myorange}\textbf{1.18} & \cellcolor{myorange}\textbf{75.44} & \cellcolor{myorange}\textbf{0.58}  \\
			4 & 500 & 40.19 \color{mygreen}{(\textbf{+0.03})} & 1.58 \color{mywarning}{(\textbf{+0.40})} & 75.53 \color{mygreen}{(\textbf{+0.09})} & 0.72 \color{mywarning}{(\textbf{+0.14})}  \\
			\hline
		\end{tabular}
	}
	\captionsetup{font=small}
	\caption{\small\textbf{Impact of pixel triplet number} (\S\ref{sc:lim}) on Mapillary Vistas $2.0$~\cite{neuhold2017mapillary} \texttt{val} and PASCAL-Person-Part~\cite{xia2017joint} \texttt{test}. \textbf{Time} indicates training time (second) for each batch.}
	\label{table:trip}
	\vspace{-10pt}
\end{table}

\clearpage
\begin{figure*}[t]
    \renewcommand\thefigure{A.1}
	\vspace{-9pt}
	\begin{center}
		\includegraphics[width=\linewidth]{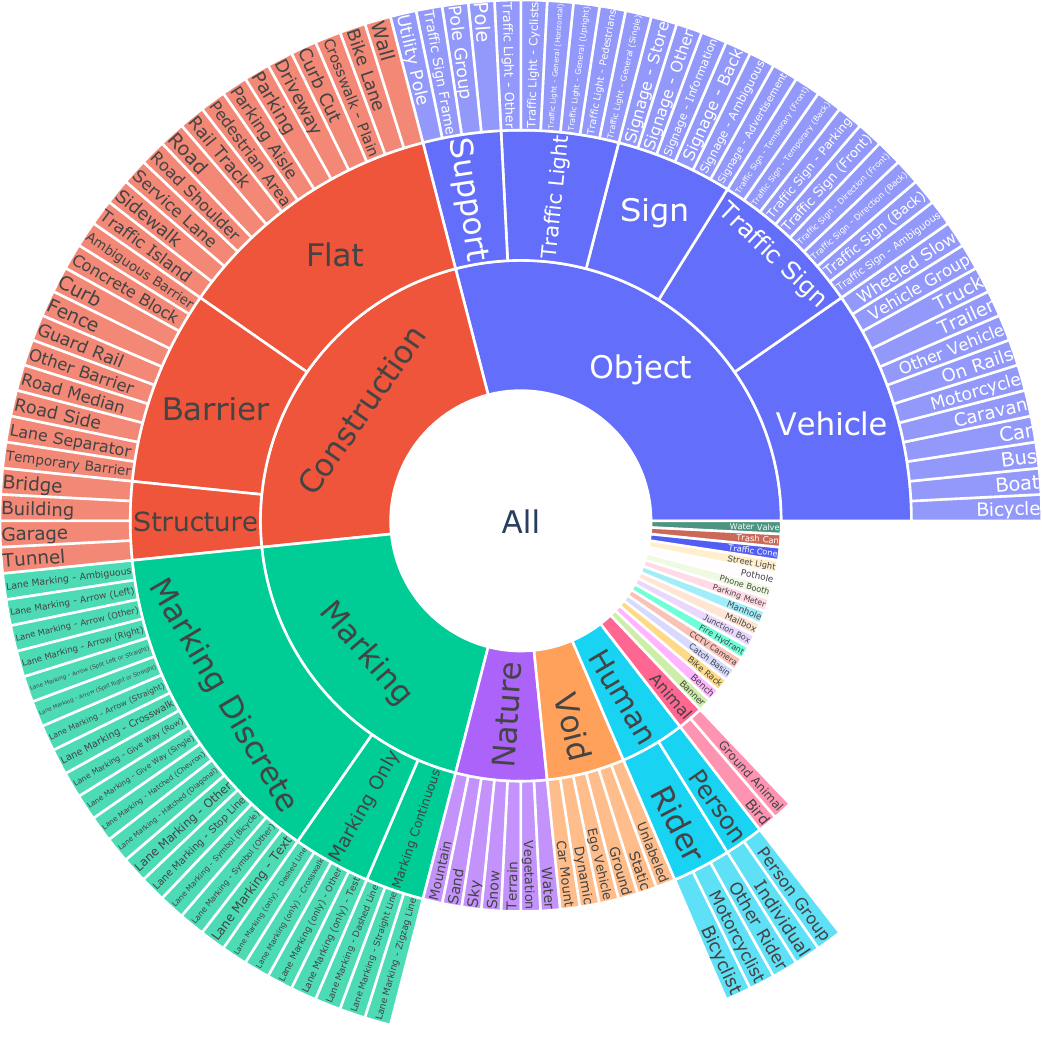}
		\end{center}
	\vspace{-17pt}
	\captionsetup{font=small}
	\caption{\small\textbf{Hierarchical architecture} of Mapillary Vistas 2.0\!~\cite{neuhold2017mapillary}.}
	\label{fig:mapillary_h}
	\vspace{-14pt}
\end{figure*}

\begin{figure}[t]
    \renewcommand\thefigure{A.2}
	\vspace{-9pt}
	\begin{center}
		\includegraphics[width=\linewidth]{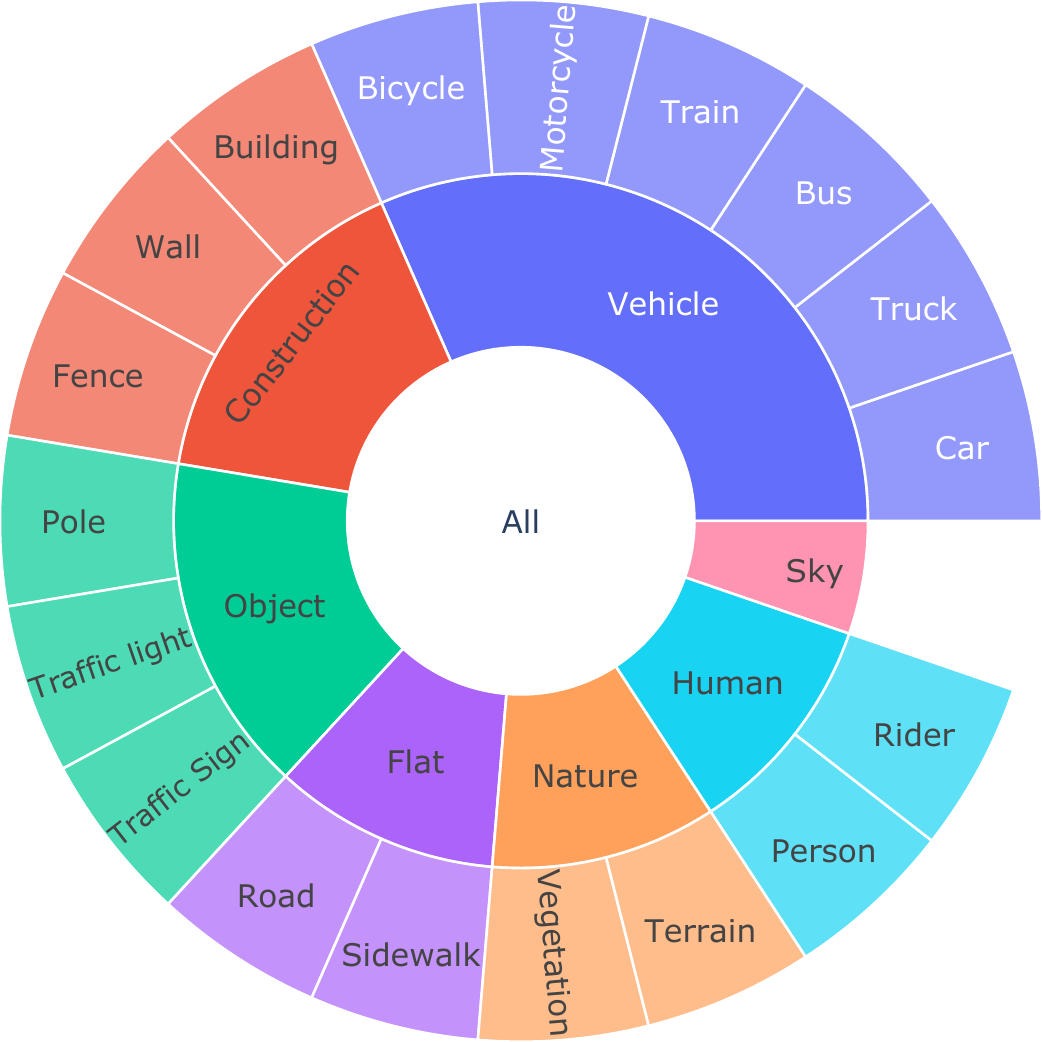}
		\end{center}
	\vspace{-17pt}
	\captionsetup{font=small}
	\caption{\small\textbf{Hierarchical architecture} of Cityscapes\!~\cite{cordts2016cityscapes}.}
	\label{fig:cityscape_h}
	\vspace{-14pt}
\end{figure}

\begin{figure}[t]
    \renewcommand\thefigure{A.3}
	\vspace{-9pt}
	\begin{center}
		\includegraphics[width=\linewidth]{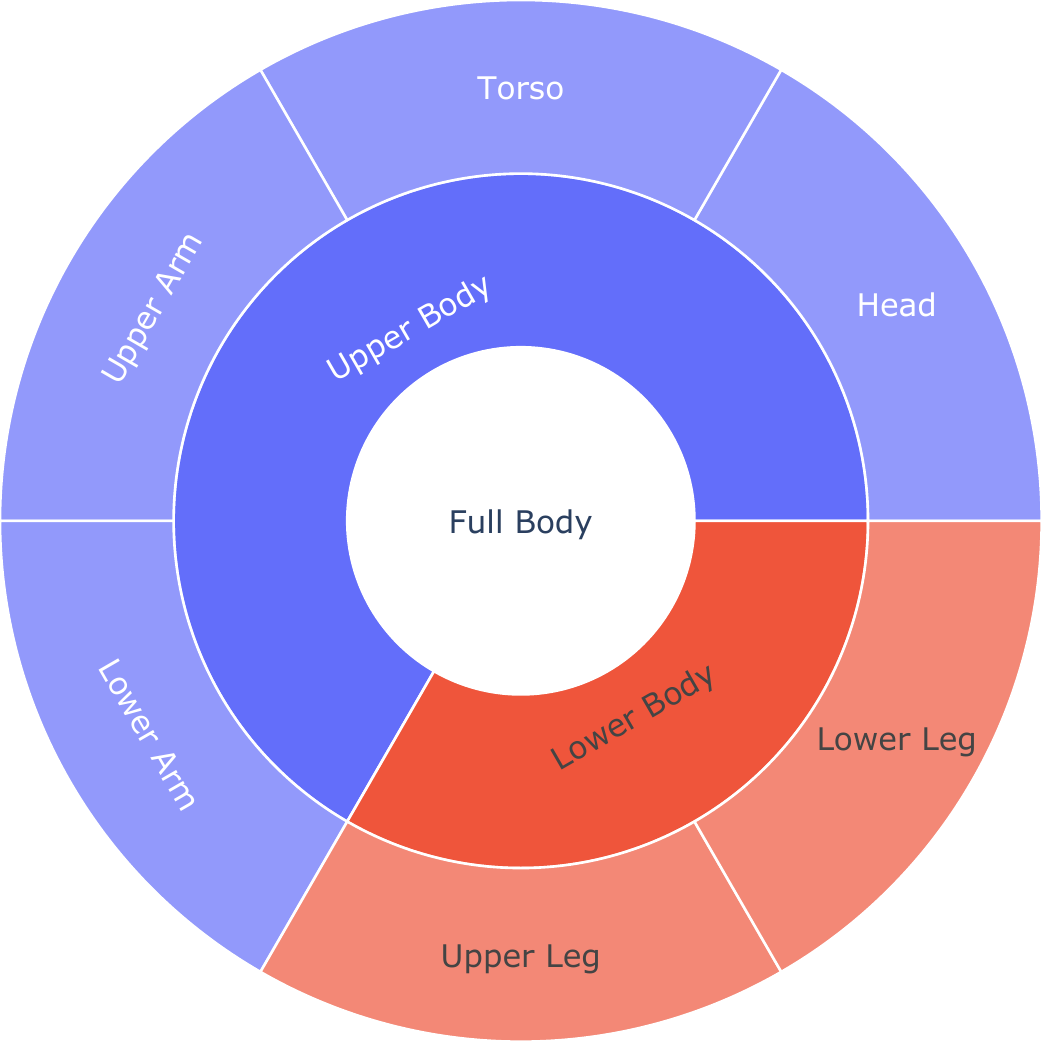}
		\end{center}
	\vspace{-17pt}
	\captionsetup{font=small}
	\caption{\small\textbf{Hierarchical architecture} of PASCAL-Person-Part\!~\cite{xia2017joint}.}
	\label{fig:ppp_h}
	\vspace{-14pt}
\end{figure}

\begin{figure}[t]
    \renewcommand\thefigure{A.4}
	\vspace{-9pt}
	\begin{center}
		\includegraphics[width=\linewidth]{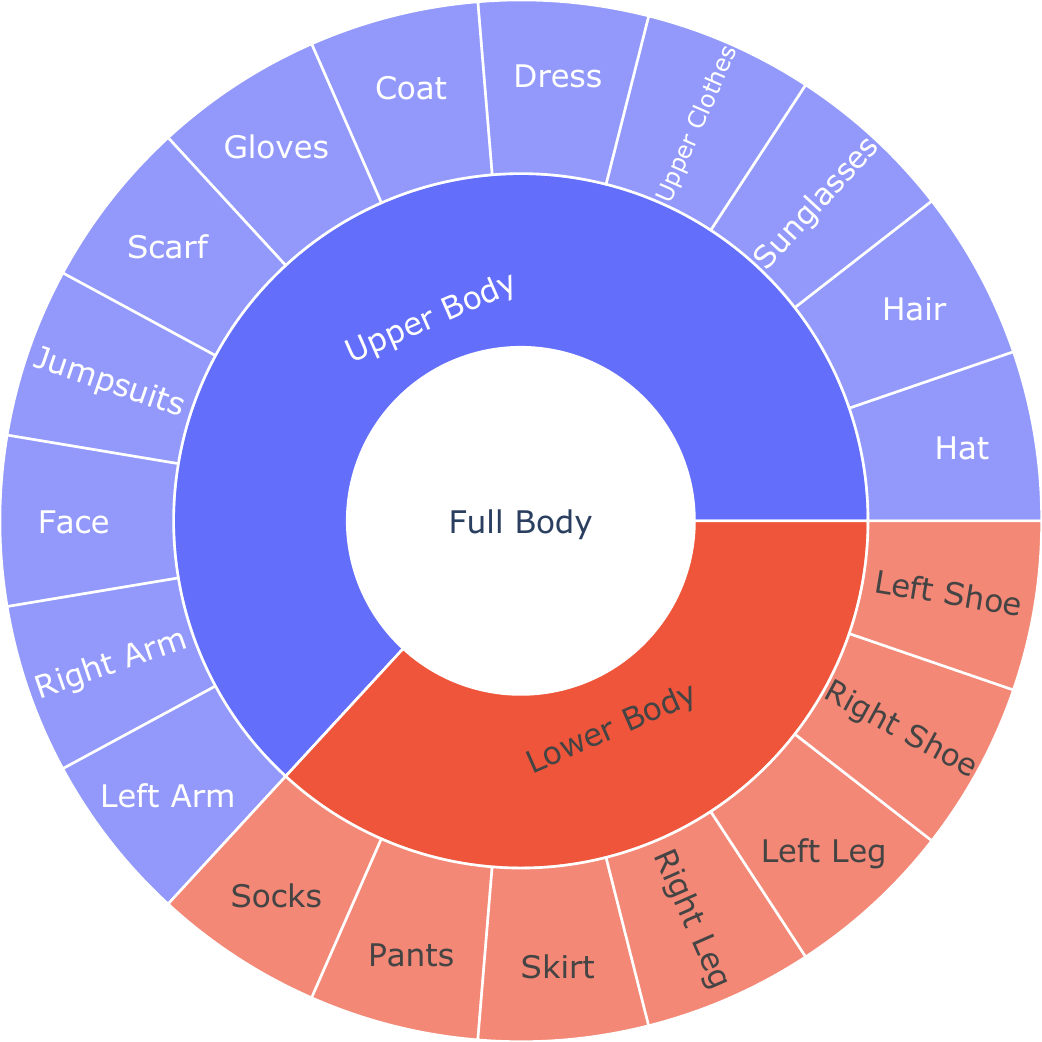}
		\end{center}
	\vspace{-17pt}
	\captionsetup{font=small}
	\caption{\small\textbf{Hierarchical architecture} of LIP\!~\cite{liang2018look}.}
	\label{fig:lip_h}
	\vspace{-14pt}
\end{figure}

\begin{figure*}[t]
    \renewcommand\thefigure{B.1}
	\vspace{-9pt}
	\begin{center}
		\includegraphics[width=\linewidth]{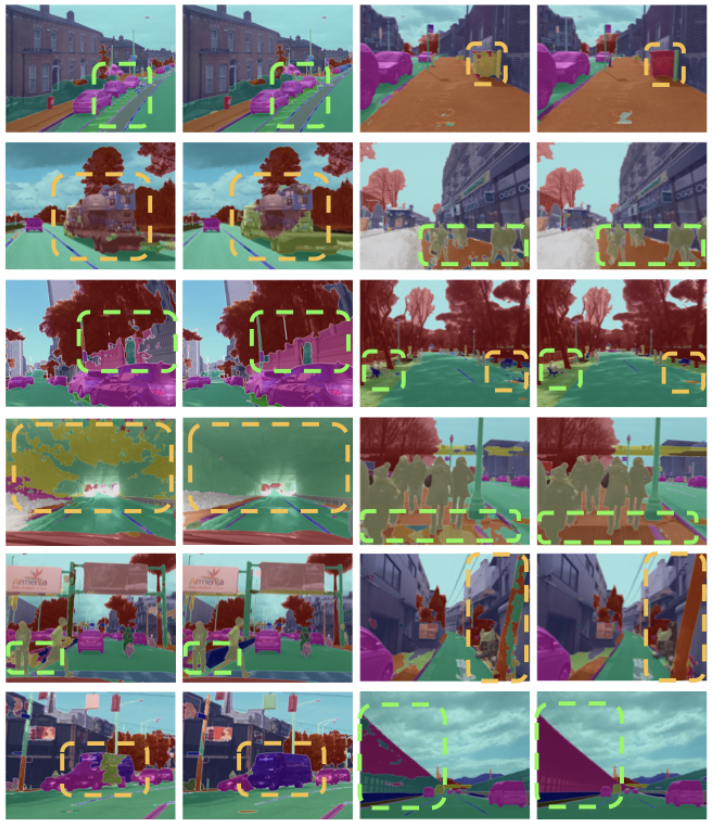}
		\end{center}
	\vspace{-17pt}
	\captionsetup{font=small}
	\caption{\small\textbf{More visualization results for semantic segmentation} on Mapillary Vistas 2.0\!~\cite{neuhold2017mapillary} \texttt{val}.}
	\label{fig:mapillary}
	\vspace{-14pt}
\end{figure*}

\begin{figure*}[t]
    \renewcommand\thefigure{B.2}
	\vspace{-9pt}
	\begin{center}
		\includegraphics[width=\linewidth]{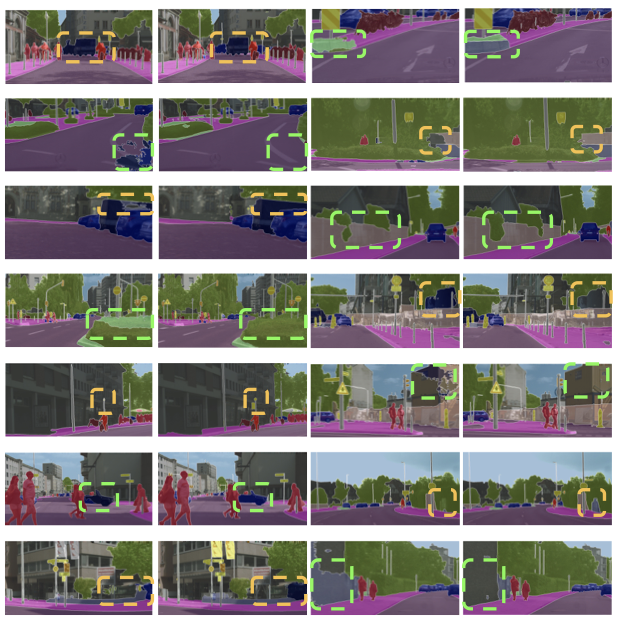}
		\end{center}
	\vspace{-17pt}
	\captionsetup{font=small}
	\caption{\small\textbf{More visualization results for semantic segmentation} on Cityscapes\!~\cite{cordts2016cityscapes} \texttt{val}.}
	\vspace{-14pt}
	\label{fig:cityscape}
\end{figure*}

\begin{figure*}[t]
    \renewcommand\thefigure{B.3}
	\vspace{-9pt}
	\begin{center}
		\includegraphics[width=\linewidth]{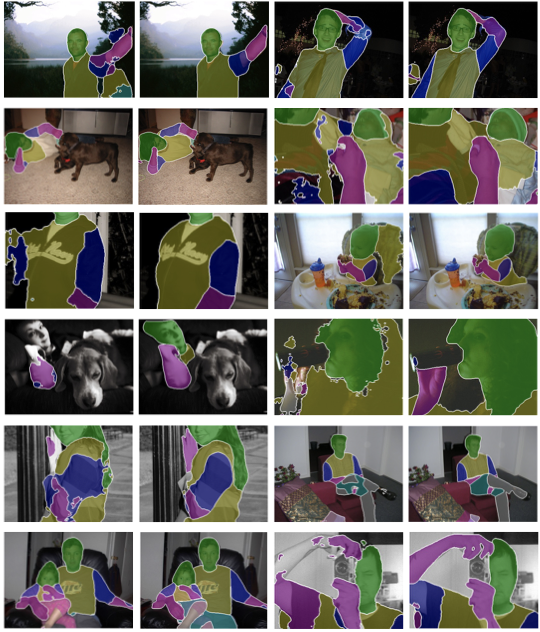}
		\end{center}
	\vspace{-17pt}
	\captionsetup{font=small}
	\caption{\small\textbf{More visualization results for human parsing} on PASCAL-Person-Part\!~\cite{xia2017joint} \texttt{val}.}
	\label{fig:ppp}
	\vspace{-14pt}
\end{figure*}

\begin{figure*}[t]
    \renewcommand\thefigure{B.4}
	\vspace{-9pt}
	\begin{center}
		\includegraphics[width=\linewidth]{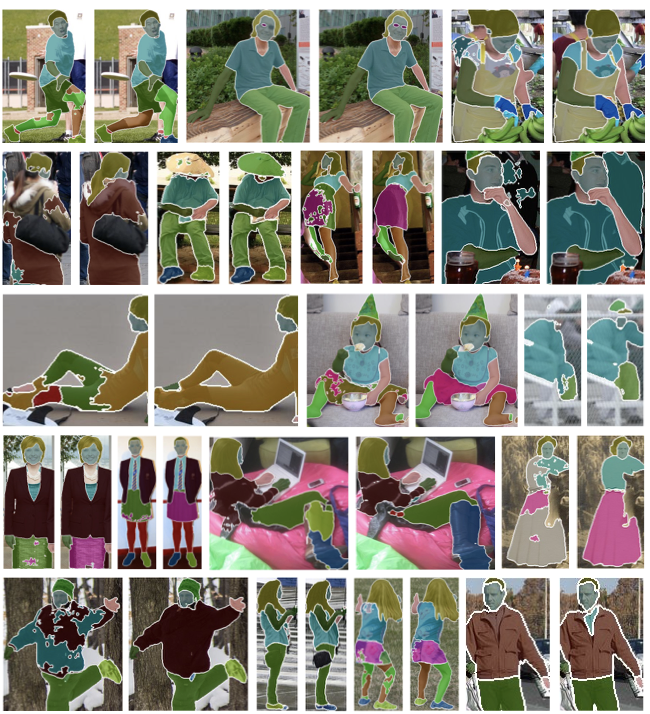}
		\end{center}
	\vspace{-17pt}
	\captionsetup{font=small}
	\caption{\small\textbf{More visualization results for human parsing} on LIP\!~\cite{liang2018look} \texttt{val}.}
	\vspace{-14pt}
	\label{fig:lip}
\end{figure*}

\end{document}